\documentclass[conference]{IEEEtran}
\IEEEoverridecommandlockouts
\usepackage{cite}
\usepackage{amsmath,amssymb,amsfonts}
\usepackage{graphicx}
\usepackage{textcomp}
\usepackage{xcolor}
\usepackage{multirow}
\usepackage{RobStd}
\usepackage{multirow}
\usepackage{booktabs}
\usepackage{enumitem}
\def\BibTeX{{\rm B\kern-.05em{\sc i\kern-.025em b}\kern-.08em
    T\kern-.1667em\lower.7ex\hbox{E}\kern-.125emX}}
    
\IEEEoverridecommandlockouts
\IEEEpubid{\makebox[\columnwidth]{} \hspace{\columnsep}\makebox[\columnwidth]{ }}

\usepackage{booktabs}
\usepackage{multirow}
\usepackage[noend]{algpseudocode}
\usepackage{algorithm}
\usepackage{subfigure}
\usepackage{enumitem}
\usepackage{amsthm}
\usepackage{makecell}
\newtheorem{Definition}{Definition}
\usepackage{authblk}
\usepackage{amsmath}

\begin{document}
\title{AdaptCL: Efficient Collaborative Learning with Dynamic and Adaptive Pruning}

\author[*]{Guangmeng Zhou}
\author[*]{Ke Xu}
\author[*]{Qi Li}
\author[*]{Yang Liu}
\author[*]{Yi Zhao}
\affil[*]{Department of Computer Science and Technology, Tsinghua University, Beijing, China}

\maketitle

\begin{abstract}
In multi-party collaborative learning, the parameter server sends a global model to each data holder for local training and then aggregates committed models globally to achieve privacy protection. However, both the dragger issue of synchronous collaborative learning and the staleness issue of asynchronous collaborative learning make collaborative learning inefficient in real-world heterogeneous environments.
We propose a novel and efficient collaborative learning framework named AdaptCL, which generates an adaptive sub-model dynamically from the global base model for each data holder, without any prior information about worker capability. All workers (data holders) achieve approximately identical update time as the fastest worker by equipping them with capability-adapted pruned models. Thus the training process can be dramatically accelerated. Besides, we tailor the efficient pruned rate learning algorithm and pruning approach for AdaptCL. Meanwhile, AdaptCL provides a mechanism for handling the trade-off between accuracy and time overhead and can be combined with other techniques to accelerate training further.
Empirical results show that AdaptCL introduces little computing and communication overhead. AdaptCL achieves time savings of more than 41\% on average and improves accuracy in a low heterogeneous environment. In a highly heterogeneous environment, AdaptCL achieves a training speedup of 6.2x with a slight loss of accuracy.
\end{abstract}

\section{Introduction}

Deep learning has made great progress in many fields (e.g., computer vision and natural language processing). It is generally agreed that the effect of the model is closely related to the amount of data used for training. However, due to data privacy or the high cost of data migration (e.g., face recognition data, mobile phone behavior data, and cross-domain network traffic data), data cannot be collected centrally for training in some scenes. As a new machine learning paradigm to solve this problem, collaborative learning \cite{shokri2015privacy} has received considerable attention. Instead of putting data to the model location, collaborative learning pushes the model to the data location. In collaborative learning, 
the training process of each round includes the following steps:

\begin{enumerate}[itemsep= 2 pt,topsep = 2 pt]
\item The server sends the current global model to workers.
\item Workers perform local updates and send the updated model to the server.
\item The server aggregates updated models from workers.
\end{enumerate}

In practical applications, workers of collaborative learning are typically resource-constrained and thus heterogeneous in computing and communication capabilities (e.g., edge devices are equipped with different computing chips and located in different domains). Even with the same physical equipment, the resource (e.g., the memory and bandwidth) allocated to a task is usually limited. Besides, the capability of a worker may fluctuate over time (e.g., a user's phone may have higher bandwidth to transmit model updates at night). In \cite{shokri2015privacy}, the asynchronous parallel (ASP) policy that the server updates the global model as soon as it receives an update is applied. The asynchronous approach improves system throughput but also introduces the staleness issue in heterogeneous environments. The staleness issue means that the model trained at the slowest worker is many rounds behind the latest global model, and the updates obtained from the slowest worker may damage the latest global model or even lead to the dilemma of non-convergence \cite{dai2018toward}. \cite{mcmahan2017communication} developed a framework named federated learning and reported more empirical results, extending collaborative learning to numerous smart device collaboration.  In federated learning, the bulk synchronous parallel (BSP) policy \cite{valiant1990bridging} that the server updates the global model until updates of all workers or at least a certain number of workers have committed is applied. Given the heterogeneity of workers, the slowest worker in collaboration drags down the entire training process. Therefore, both of the above frameworks are inefficient in real heterogeneous environments. We assert that \textbf{the root cause of the inefficiency in heterogeneous environments is different capabilities workers are required to do the same thing (e.g., training the same model)}. 

Thus, we claim that different capabilities workers should train models of different sizes, pruned from the global base model, to achieve approximately identical update time per round. Over the years, many network pruning studies claim they can prune 60-70\% of the parameters in a pre-trained neural network with little or no loss of accuracy after fine-tune \cite{Tan2020DropNetRN,lin2020hrank}. However, directly applying these network pruning techniques is not suitable, i.e., they are based on pre-trained models to speed up inference as opposed to our intention to speed up training and based on the model being trained. Considering there is a gap between what we do and what we pursue in distributed pruning, i.e., we prune the sub-model but pursue the accuracy of the global model, the effectiveness of previous techniques remains to be explored. Besides, we expect pruning and fine-tune to be extremely time-efficient. In addition to pruning, designing precise and adaptive per-round pruned rates so that all workers achieve approximately identical update time quickly is challenging, especially in a dynamic environment.

In this paper, we propose a novel and efficient synchronous collaborative learning framework named {\bf AdaptCL}, which generates an adaptive sparse sub-model dynamically from the global base model for each data holder, based on their computing and communication capabilities. By equipping different capabilities workers with adaptive size models, worker update time (including local training and communication time) is adaptively adjusted. Thus all workers achieve approximately identical update time as the fastest worker. In this way, AdaptCL achieves over-asynchronous efficiency using the synchronous approach and avoid the staleness issue.

Specifically, the server takes the worker's update time per round to characterize its capabilities and dynamically builds personalized modeling of the worker from data collected on model retention ratio and corresponding update time. When reaching the pruning round, the server sets the adaptive pruned rate for each worker with the shortest update time as the target. Then worker prunes and reconfigures the local model according to the pruned rate.

We conduct extensive experiments on CIFAR10, CIFAR100, and Tiny-ImageNet with VGG16 and ResNet50 models and take Non-IID (independent and identically distributed) cases into account. Compared with synchronous and asynchronous federated learning, AdaptCL demonstrates its efficiency. AdaptCL achieves time savings of more than 41\% on average in a low heterogeneous environment and improves accuracy. In a highly heterogeneous environment, AdaptCL achieves a training speedup of 6.2x with a slight loss of accuracy. Also, AdaptCL provides controlling parameters that can do accuracy and training time trade-offs. Besides, AdaptCL can be combined with other approaches to achieve higher speedups.

To summarize, we make the following contributions:
\begin{itemize}[itemsep= 0.5 pt, topsep = 0.5 pt]

\item To the best of our knowledge, we are the first to leverage dynamic and adaptive sub-models to solve the dragger issue of synchronization caused by the worker capacity heterogeneity.

\item We propose an efficient collaborative learning framework {\it AdaptCL}, which performs distributed pruning for participating workers to speed up the entire training process while maintaining satisfactory accuracy.

\item We propose a dynamic and adaptive pruned rate learning method that enables each worker's update time to quickly achieve identical without any prior capability-related information.

\item We discover and understand the pruning principles in distributed pruning and tailor a novel and efficient pruning method \emph{CIG-BNscalor} for distributed pruning.

\item Extensive experiments show that AdaptCL exhibits efficiency in various data-model settings and different degrees of heterogeneous environments.


\end{itemize}
\section{Related Work}

\subsection{Network Pruning} \label{section:network prune}
Network pruning is based on the premise that deep neural networks are over-parameterized, and thus proper pruning can cut off a large number of parameters to
accelerate model inference while maintaining accuracy. Network pruning consists of three stages: training, pruning, and fine-tune. According to the pruned object, network pruning can be divided into non-structural pruning, which only cuts off weights, and structural pruning, which cuts off units such as neurons and filters. However, non-structural pruning can only achieve inference
acceleration on specialized software \cite{park2016faster} or hardware \cite{han2016eie}, while structural pruning has no such limitations and thus it is getting more attention. 

There are two critical points with structural pruning: how to prune and how much to prune. For the first one, structural pruning usually cuts off units of low importance predefined, such as the percentage of zero activation \cite{hu2016network}, the $\ell1$-norm \cite{li2016pruning}, the mean of product of first-order gradient and weight \cite{molchanov2016pruning}, the scaling factor of Batch normalization (BN) layer \cite{liu2017learning}, the distance from the geometric median of filters of same layer \cite{he2019filter}, the rank of feature map \cite{lin2020hrank}, or the importance indicators introduced in training \cite{huang2018data,lin2019towards,you2019gate}. For the second one, different network layers express different semantics \cite{bau2020understanding} and have different sensitivities to parameter pruning \cite{han2015learning}, so the pruned rate per layer should be different for a given pruning budget. \cite{chin2020towards,molchanov2016pruning,you2019gate} try to get global rank for all filters, and \cite{kusupati2020soft} proposes to learn pruning thresholds for different layers. Besides, there is some research concerned with changing prune from unidirectional to dynamic. \cite{he2018soft,lin2020dynamic} continue updating the pruned filters during training so that they can be reactivated later. 

However, directly applying these techniques is not suitable—they are based on pre-trained models to speed up inference as opposed to our intention to speed up training and based on the model being trained. While some ideas could be adapted, we expect the pruning and fine-tune to be extremely time-efficient and can be done early in the training process, which is not considered in previous works.

\subsection{Efficient Collaborative Learning}
\label{subsection:Efficient Collaborative Learning}

Collaborative learning, a new machine learning paradigm, is closely related to distributed machine learning. However, the relatively lower level of trust and a higher degree of heterogeneity between workers make collaborative learning more problematic in terms of security, privacy, efficiency, optimization, etc. We focus on the fundamental issue, i.e., efficiency in this paper and divide the causes of inefficient training into local and global causes. The local cause is that the model is too large, leading to extremely time-consuming model training and transmission. The global cause is the heterogeneity between workers, leading to the presence of draggers.
 
 {\bf Local cause}. To speed up model transmission, gradient quantization and sparsification are extensively studied. Gradient quantization is achieved by quantizing the gradients to low-precision values \cite{wen2017terngrad}. Gradient sparsification is usually done by selecting significant parameters to transmit \cite{hsieh2017gaia,konevcny2016federated,lin2018deep} or adding constraints to get a sparse model \cite{caldas2018expanding}. In addition to reducing transmitted parameters to speed up single transmission, efficiency can also be improved by adjusting the frequency of parameter aggregation \cite{wang2019adaptive}. In contrast, little research has been done on speed up training. Adopting lightweight networks may be a viable approach, such as MobileNet \cite{howard2017mobilenets} and ShuffleNet \cite{ma2018shufflenet,zhang2018shufflenet}. Network pruning can speed up model transmission, but almost all network pruning studies cannot speed up training because they retain the original dense model during pruning. \cite{lym2019prunetrain} solve the problem by reconfiguring the model into smaller models during pruning. We draw on the idea of reconfiguration to speed up, but we are different in three aspects: 1) AdaptCL is a distributed pruning framework, pursuing global model accuracy. 2) Our prune budget is inter-workers varying and dynamic, derived from an external factor of training, i.e., worker update time, not on the parameter's threshold. Thus we are more challenging to recover model accuracy. 3)We consider the Non-IID data problem in distributed training. 

{\bf Global cause}. Except for BSP and ASP, other server synchronization policies have been extensively studied. The stale synchronous parallel (SSP) policy \cite{dai2015high,ho2013more} is a trade-off between BSP and ASP, in which when the fastest worker is ahead of the slowest worker by predefined threshold $s$ rounds, the fastest worker needs to stop and wait. \cite{zhu2019learning} develops a unified synchronization policy framework that can cover BSP, ASP, and SSP by designing the collection of active workers of the next round, and then uses reinforcement learning to learn the collection to minimize the total time cost. However, these ways are still doing trade-off between BSP and ASP. Instead of doing a trade-off, we start with BSP and solve the dragger issue by giving workers different size models to make update time close to the same.

Before elaborating on our framework, we summarize the main notations in this paper in Tab. \ref{table:notations}.

\begin{table}[t]
\begin{center}
\begin{tabular}{lp{0.8\linewidth}l}
\toprule
$T$ & Number of communication rounds \\
$W$ & Number of workers \\
$E$ & Number of local training epochs \\
$H$ & Heterogeneity \\
$D_{w}$ & Data of worker $w$ \\
$\theta_{g}^{t}$ & Parameters of global model at round t \\
$\theta_{w}^{t}$ & Parameters of worker $w$'s model at round t \\
$I_{w}^{t}$ & Unit indexes of worker $w$'s sub-model corresponding to the global model at round t \\
${P}_{w}^{t}$ & Pruned rate of worker $w$ at round t \\

${\phi}_{w}^{t}$ & Update time of worker $w$ at round t \\

$PI$ & Pruned interval \\
$\beta$ & Ratio of the first training part \\
$\gamma$ & Model retention ratio \\
${\gamma}_{min}$ & Minimum model retention ratio set for pruning \\
${\rho}_{min}$ & Minimum pruned rate set for pruning \\ 
${\rho}_{max}$ & Maximum pruned rate set for pruning \\
$\alpha$ & Initial coefficients on update time and model retention ratio set for pruning \\
$\sigma$ & Ratio of the longest update time to the shortest update time set in Exp (experiment) \\
$B_{max}$ & Bandwidth of the fastest worker set in Exp \\
\bottomrule
\end{tabular}
\end{center}
\caption{Summary of main notations.}
\label{table:notations}
\end{table}

\begin{figure}[t]
\begin{center}
  \includegraphics[width=1.0\linewidth]{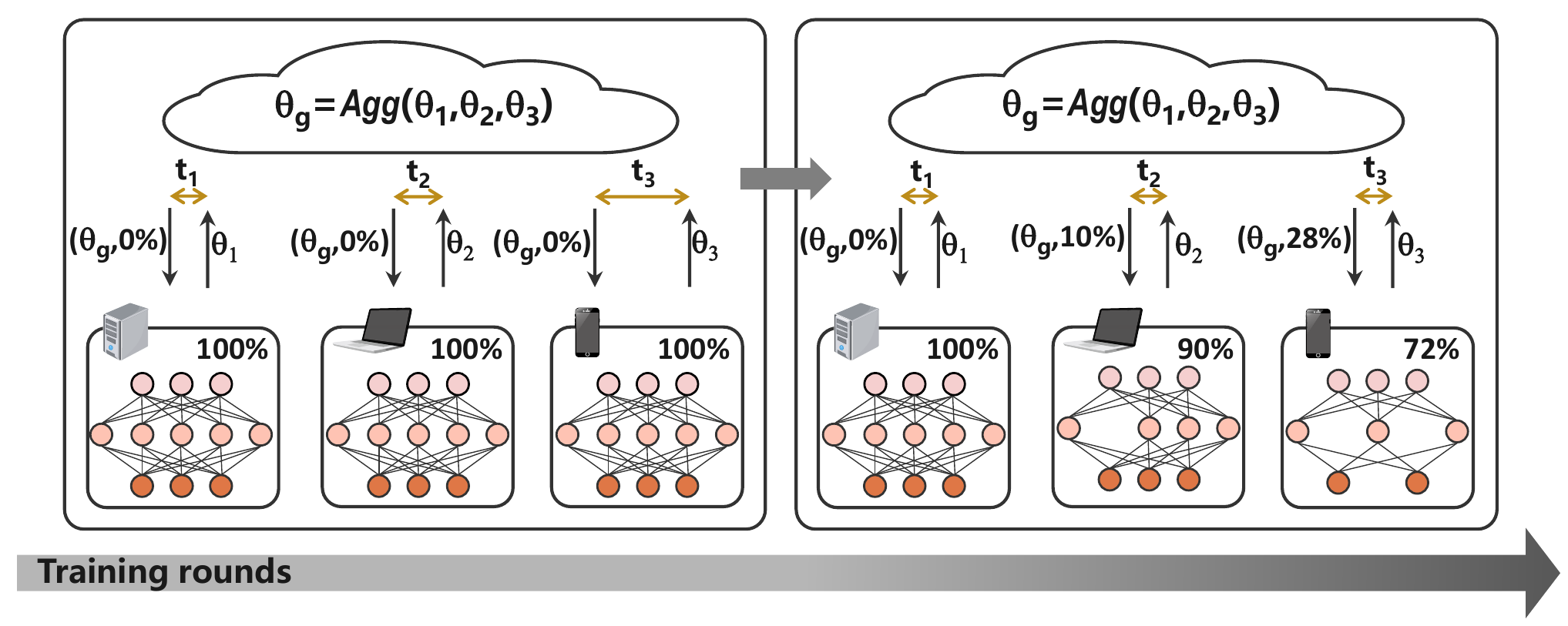}
\end{center}
  \caption{AdaptCL process: the cloud server keeps issuing pruned rates to resize workers' models during training until all workers have the same update time. $t_{1}$, $t_{2}$ and $t_{3}$ represent the update time of the above three workers, respectively.}
\label{figure:process}
\end{figure}

\section{AdaptCL}
In this section, we overview our framework AdaptCL firstly, then elaborate on the critical components of the framework, including how to do model training and aggregating, how to prune, and how to determine pruned rates derived from worker update time.

\subsection{Overview}

\begin{algorithm}[t] 
\caption{AdaptCL: Collaborative Learning with Dynamic and Adaptive Pruning} 
\label{algorithm:framework}
\hspace*{0.02in} {\bf Server:}
\begin{algorithmic}[1]

\For{each round $t=1,...,T$}
\While{not receiving all updates}
\State Server receives $\theta_{w}^{t}$, $I_{w}^{t}$ from worker $w$, \\ \qquad \quad and calculate worker $w$' update time ${\phi}_{w}^{t}$ 
\EndWhile
\State $\theta_{g}^{t}$ $\gets$ ModelAggregate($\theta_{1}^{t}$, $\theta_{2}^{t}$,..., $\theta_{W}^{t}$)

\If{it is pruning round}
\State Obtain \{${P}_{1}^{t+1},...,{P}_{W}^{t+1}$\} with Algorithm \ref{algorithm:GetPrunedRate}
\EndIf
\For {each worker $w=1,...,W$}
\State $\theta_{w}^{t+1}$ $\gets$ $\theta_{g}^{t} \odot I_{w}^{t}$
\State Server send $\theta_{w}^{t+1}$ and ${P}_{w}^{t+1}$ to worker $w$
\EndFor
\EndFor
\end{algorithmic}

\hspace*{0.02in} {\bf Worker:} 
\begin{algorithmic}[1]

\State Worker $w$ receives $\theta_{w}^{t+1}$ and ${P}_{w}^{t+1}$ from server

\State $\theta_{w}^{t+1}$ $\gets$ SparseTrain($ \beta E, \theta_{w}^{t+1}, D_{w}$)
\If{${P}_{w}^{t+1} >$ 0}

\State $mask$ $\gets$ NetworkPrune(${P}_{w}^{t+1}$)
\State $\theta_{w}^{t+1}$, $I_{w}^{t+1}$ $\gets$ NetworkReconfigure($\theta_{w}^{t+1}$, $mask$)

\Else
\State $I_{w}^{t+1}$ $\gets$ $I_{w}^{t}$
\EndIf
\State $\theta_{w}^{t+1}$ $\gets$ SparseTrain($ (1-\beta) E,\theta_{w}^{t+1}, D_{w}$)
\State Worker $w$ send $\theta_{w}^{t+1}$ and $I_{w}^{t+1}$ to server
\end{algorithmic}
\end{algorithm}

The process of AdaptCL is shown in Fig. \ref{figure:process}. When training begins, the cloud parameter server distributes the same model to each worker conservatively since workers' capabilities are unknown. After obtaining the worker capability signal, i.e., the update time, the server generates an adaptive pruned rate for each worker through the designed pruned rate learning algorithm. The worker receives the pruned rate and then prunes its model locally. The whole process proceeds dynamically with training until the update time tends to identical.

The pseudo-code of the framework is shown in Algorithm \ref{algorithm:framework}. AdaptCL consists of two types of participants, server and worker. On the server-side, we adopt the synchronous approach that the server starts aggregating until all workers' updates are received as in federated learning \cite{mcmahan2017communication}. If the pruned round arrives, the pruned rate learning algorithm \ref{algorithm:GetPrunedRate} is used to get the pruned rate for each worker. Finally, the server obtains the parameters of the worker's sub-model by $\theta_{g}^{t} \odot I_{w}^{t}$, and sends the parameters and pruned rate to the worker. On the worker-side, the worker trains part of the epochs after receiving parameters. If pruning is not needed, the worker continues training the other part of the epochs on the previous model. When pruning is required, we first follow a specific pruning pattern to cut units, then build and initialize the sub-model with the previous weights, and finally update the local model's global index corresponding to the global model. In the end, the parameters and global index of the current model are sent to the server. It is worth noting that, compared to federated learning, the content of the communication is only more model's global index as well as pruned rate, which introduces little communication overhead.

\subsection{Model Training and Aggregating}
\label{section:model training}

{\bf Sparse training}. We use the same sparse training approach as in the \cite{lym2019prunetrain}. The loss function is shown in Eq. \ref{equation:loss}, consists of cross-entropy loss and group lasso loss, in which $\mathcal{B}$ represents a batch of data and $\mathcal{G}$ represents groups of parameters. With the introduction of group lasso regularization, the parameters associated with a unit are viewed as a group, and the training reduces the value of a group of parameters at the same time, thus mitigating the impact caused by pruning the unit. In addition, we divide the training process into two parts to explore the effects of different positions of pruning in training by setting different $\beta$.

\begin{equation}
    \mathop{min}\limits_{\theta} ( \frac{1}{\vert \mathcal{B} \vert} \sum_{(x,y) \in \mathcal{B}} l(y,f(x,\theta)) + \lambda \cdot \sum_{g \in \mathcal{G}} \sqrt{\vert g \vert} \left\|\theta_{g}\right\|_2 ) \label{equation:loss}
\end{equation}

{\bf Model aggregating}. As a result of network pruning, a unit may only exist in $w^{'}$ ($w^{'}<W$) sub-models, thus there are two approaches to set the aggregation coefficient of the unit when doing aggregation. One way is {\bf By-unit} with a coefficient of $\frac{1}{w^{'}}$, and the other way is {\bf By-worker}, with a coefficient of $\frac{1}{W}$, if we ignore the differences in the amount of data between workers here. The two approaches are illustrated in  Appendix \ref{appendix:AdaptCL details} Fig. \ref{figure:modelagg}. \cite{zhou2019deconstructing} points out that the reason lottery tickets behave better is that mask operation freezes some values to zero and can make those values reach the end of their optimization process faster. By-worker is equivalent to treating the parameters of a unit as $0$ in a sub-model that does not contain that unit. Therefore, we adopt by-worker to do model aggregating in AdaptCL.

\subsection{Pruned Rate Learning}

\begin{algorithm}[t] 
\caption{Pruned Rate Learning}
\label{algorithm:GetPrunedRate}
\hspace*{0.02in} {\bf Input:} 
round $t$, the model retention ratio ${\gamma}_{w}$ and corresponding average update time ${\phi}_{w}$ of previous rounds. ${\gamma}_{w}^{now}$ and ${\phi}_{w}^{now}$ represent the model retention ratio now and corresponding average update time, respectively. \\
\hspace*{0.02in} {\bf Output:}
${P}_{1}^{t+1},{P}_{2}^{t+1},...,{P}_{W}^{t+1}$
\begin{algorithmic}[1]
    \State ${\phi}_{min} \gets min({\phi}_{1}^{now},{\phi}_{2}^{now},...,{\phi}_{W}^{now})$
    \For {each worker $w=1,...,W$}
        \If{the worker has been pruned}
        \State Get ${\gamma}_{target}$ by Newton interpolation as Eq. \ref{equation:prunedrate}
          \If {${\gamma}_{now}-max({\gamma}_{target}, {\gamma}_{min}) < {\rho}_{min}$}
          \State ${\gamma}_{target}$ $\gets$ ${\gamma}_{now}$
          \EndIf
          \State ${P}_{w}^{t+1} \gets \frac{{\gamma}_{now} - {\gamma}_{target}}{{\gamma}_{now}}$
        \Else
          \State ${P}_{w}^{t+1} \gets \frac{ {\phi}_{w}^{now} - {\phi}_{min}}{\alpha*{\phi}_{w}^{now}}$
        \EndIf
        \State ${P}_{w}^{t+1} \gets min({P}_{w}^{t+1}, {\rho}_{max})$

    \EndFor
\end{algorithmic}
\end{algorithm}

We determine the pruned rate based on the worker's update time, consisting of three parts: send time, training time, and receive time. The send and receive times can be considered to vary linearly with the size of transmitted parameters if the bandwidth is stable. Many researchers use FLOPs (floating point operations) as an indicator of training speed. However, training time is affected by many practical factors (e.g., data loading speed, parallel optimization in the computing chip, underlying optimization of the computing platform). Therefore, given a pruned rate, it is challenging to know precisely the update time after pruning in the absence of prior knowledge of the worker's computing and communication capabilities. Thus an approximate, dynamic pruned rate learning algorithm is needed.

Our pruned rate learning algorithm is shown in Alg.  \ref{algorithm:GetPrunedRate}.
First, we take the current minimum update time as the target time ${\phi}_{min}$ for the next round of pruning. Then we use the data that have been accumulated, i.e., the model retention ratio and update time after each pruning (${\gamma}_{w}^{0},{\phi}_{w}^{0}$), ..., (${\gamma}_{w}^{n},{\phi}_{w}^{n}$), to construct a polynomial $\widetilde {f}$ that satisfies $\widetilde {f}_{w}({\gamma}_{w}^{i}) = {\phi}_{w}^{i}, \ i=1,...,n$. According to Main Theorem of Polynomial Interpolation \cite{sauer2006numerical}, $\widetilde {f}_{w}$ is existent and unique. There are many interpolation methods that have been studied. Among them, Newton interpolation is solid and fast, so we adopt Newton interpolation in AdaptCL. How to obtain target retention ratio ${\gamma}_{target}$ is shown in Eq. \ref{equation:prunedrate}. 

If no pruning has been done before, we assume $\phi = \alpha * {\phi}_{w}^{now}\gamma$, and the pruned rate is obtained as in line 9. Besides, we set a minimum model retention ratio ${\gamma}_{min}$, a maximum pruned rate ${\rho}_{max}$ to prevent excessive pruning, and a minimum pruned rate ${\rho}_{min}$ to prevent overly frequent minor pruning. Our algorithm does not require prior information about worker capabilities and quickly adapt to dynamically changing environments. Also, the computational overhead introduced to the server is negligible.

\begin{equation}
\setlength{\abovedisplayskip}{0.2pt}
\setlength{\belowdisplayskip}{0.2pt}
\begin{aligned}
    {\gamma}_{target} &= \widetilde {f}_{w}^{-1}({\phi}_{min}) \\
    &= \widetilde {f}_{w}^{-1}[{\phi}_{w}^{0}] + \widetilde {f}_{w}^{-1}[{\phi}_{w}^{0},{\phi}_{w}^{1}]({\phi}_{min}-{\phi}_{w}^{0}) + ... \\
    &+ \widetilde {f}_{w}^{-1}[{\phi}_{w}^{0},{\phi}_{w}^{1},...,{\phi}_{w}^{n}]({\phi}_{min}-{\phi}_{w}^{0})...({\phi}_{min}-{\phi}_{w}^{n-1})\\
    where \quad &\widetilde {f}_{w}^{-1}[{\phi}_{w}^{i}] = \widetilde {f}_{w}^{-1}({\phi}_{w}^{i})= {\gamma}_{w}^{i},  i=0,...,n\\
    \widetilde {f}_{w}^{-1}[&{\phi}_{w}^{0},...,{\phi}_{w}^{n}] =\frac{\widetilde {f}_{w}^{-1}[{\phi}_{w}^{0},...,{\phi}_{w}^{n-1}]-\widetilde {f}_{w}^{-1}[{\phi}_{w}^{1},...,{\phi}_{w}^{n}]}{{\phi}_{w}^{n}-{\phi}_{w}^{0}}
\end{aligned}
\label{equation:prunedrate}
\end{equation}

\subsection{Network Pruning}
As we mentioned in Sec. \ref{section:network prune}, there are two critical questions with structural pruning: how to prune and how much to prune. How much to prune has been answered by the Alg. \ref{algorithm:GetPrunedRate}, and how to prune is described next. 

In distributed pruning, there is a gap between what we do and what we pursue, i.e., we prune the sub-model but pursue the accuracy of the global model. So it is questionable to prune directly based on the importance of units in the sub-model. \cite{diao2020heterofl} adopts a pruning approach that is not dependent on the sub-model, i.e., pruning in the order of unit index (called \emph{Index} later), and achieves good performance, but does not give an explanation. We attempt to understand the underlying reasons for its good performance and find a better pruning approach. We infer that good performance may be related to the fact that the pruned units are adjacent, or the pruning order is identical across all workers or constant over time. To investigate, we keep all other treatments the same and perform a number of variants as follows:

\begin{itemize}[topsep=0.5pt, itemsep= 0.1 pt]
\item \emph{No adjacent}: generate a random order and keep it \emph{identical} (all workers share the order) and \emph{constant} (all rounds share the order).
\item \emph{No identical}: select a random index start position and keep it \emph{constant} for each worker.
\item \emph{No constant}: reselect the identical index start position for all workers at each pruning. 
\end{itemize}

\begin{figure*}[htbp]
    \centering
    \subfigure[\emph{Index}, IID]{
        \label{figure:explore_index_methods_iid}
        \includegraphics[width=.18\linewidth]{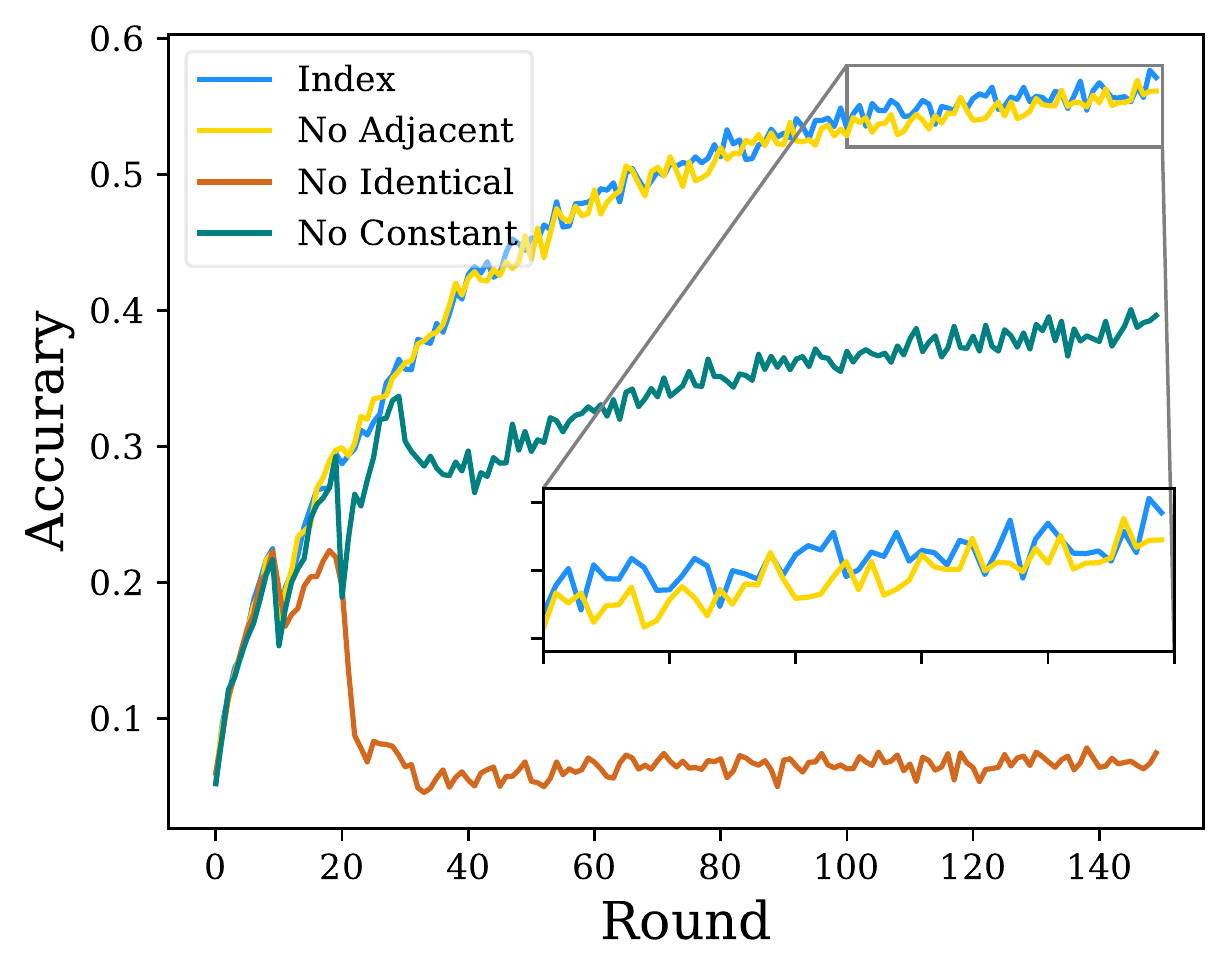}
    }
    \subfigure[\emph{Index}, Non-IID]{
    \label{figure:explore_index_methods_iid8}
	\includegraphics[width=.18\linewidth]{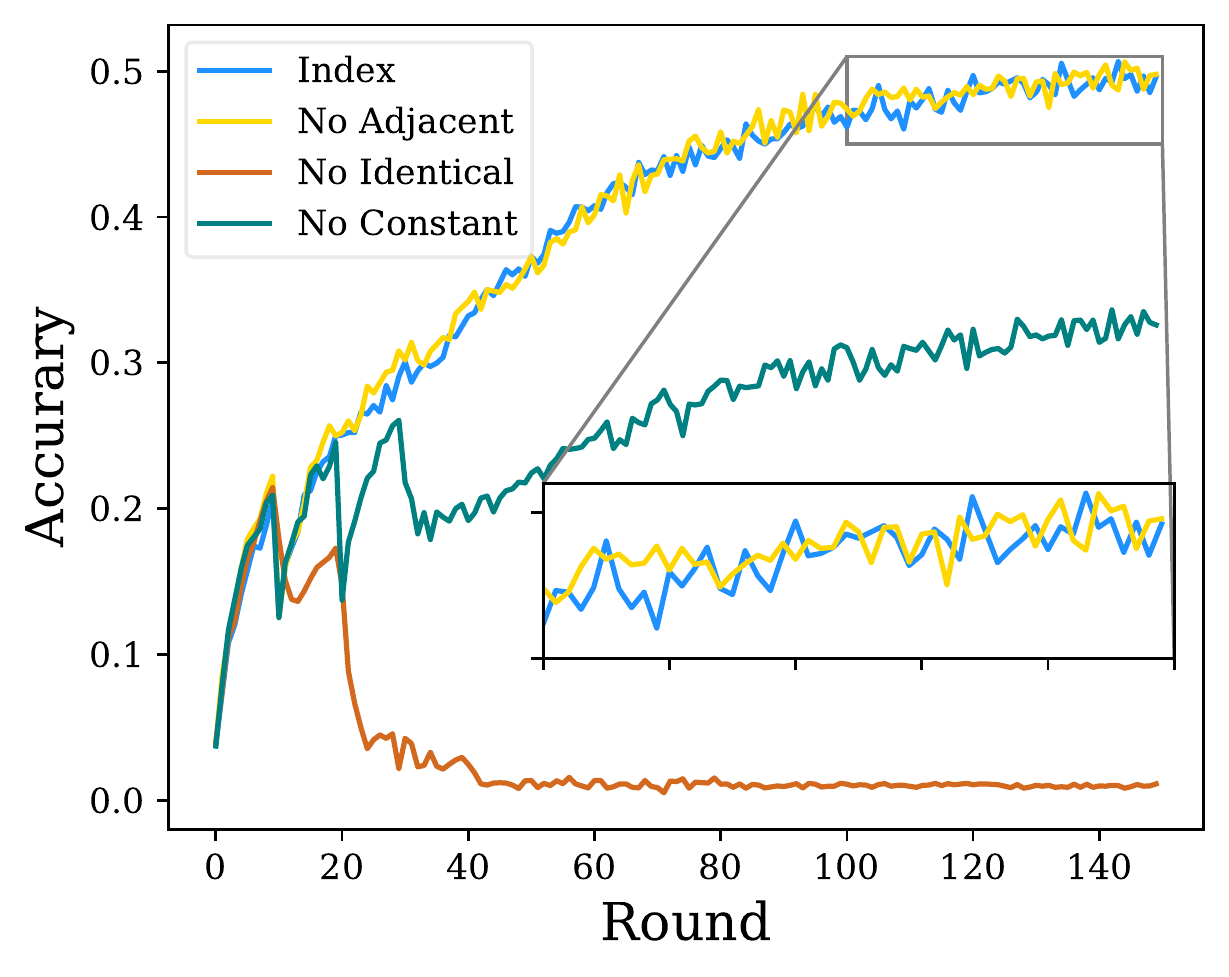}
    }
        \subfigure[Remaining Networks]{
    \label{figure:diff_method_saved_model}
	\includegraphics[width=.18\linewidth]{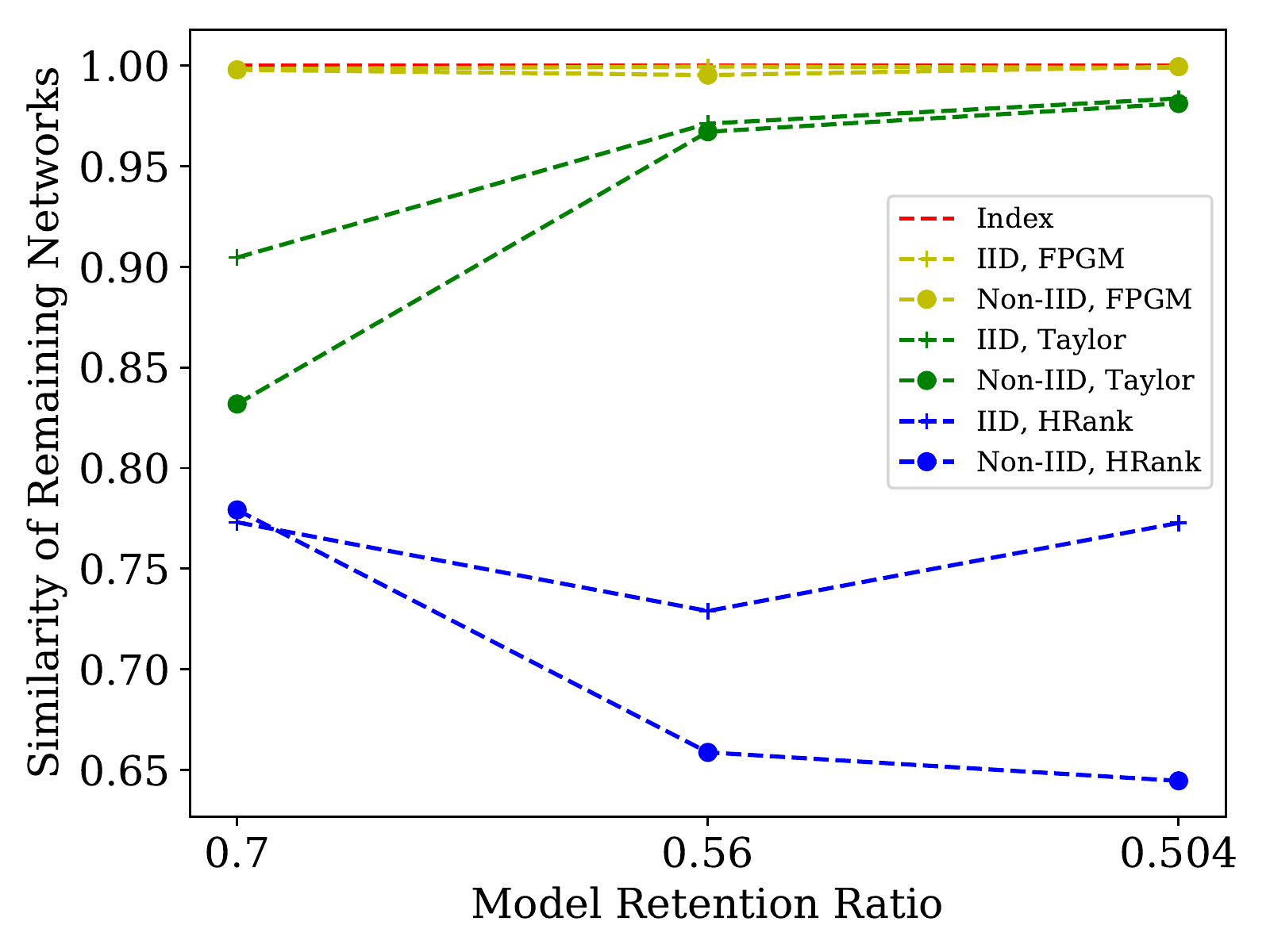}
    }
    \subfigure[Other Methods, IID]{
    \label{figure:diff_pruning_methods_iid}
	\includegraphics[width=.18\linewidth]{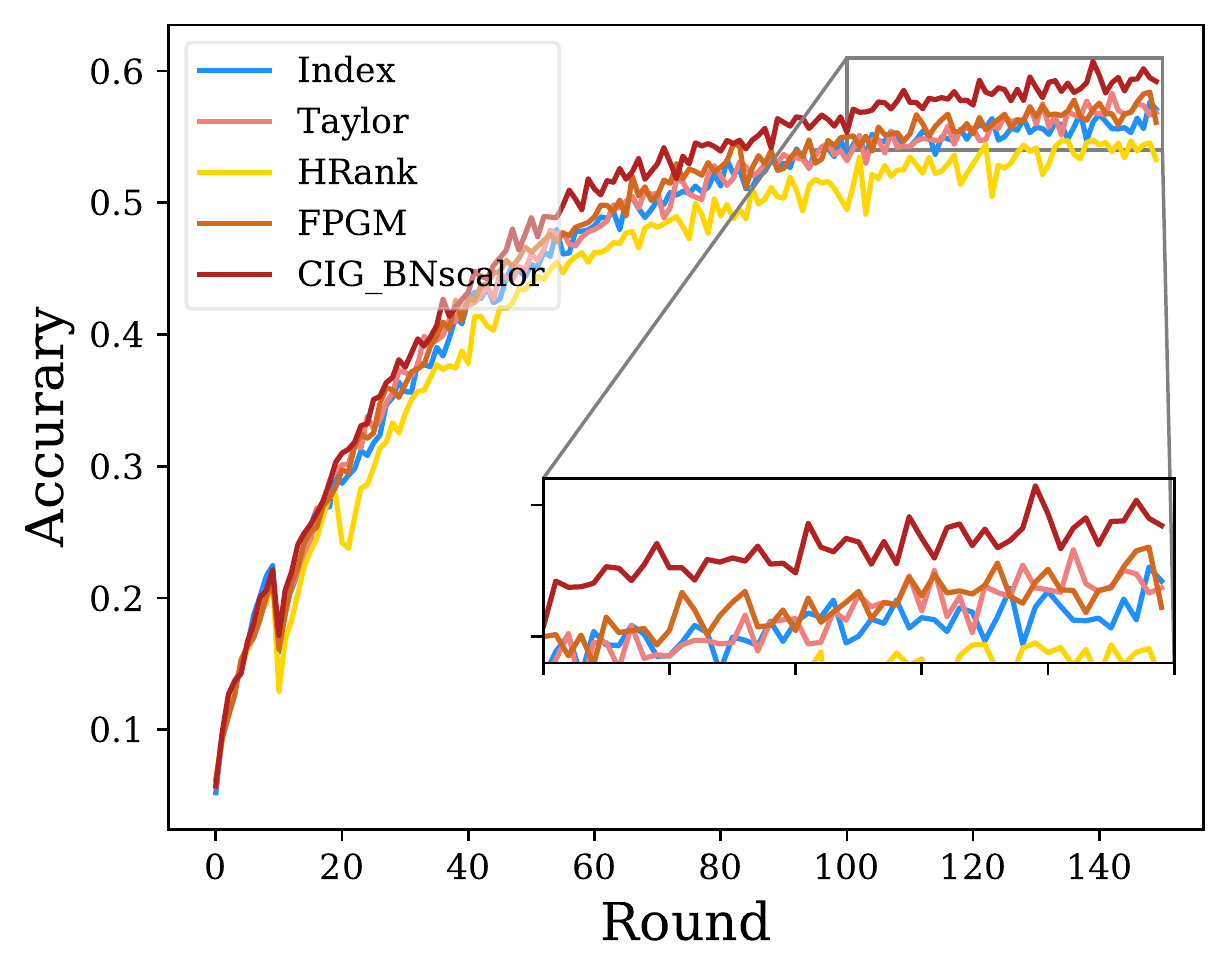}
    }
    \subfigure[Other Methods, Non-IID]{
    \label{figure:diff_pruning_methods_iid8}
	\includegraphics[width=.18\linewidth]{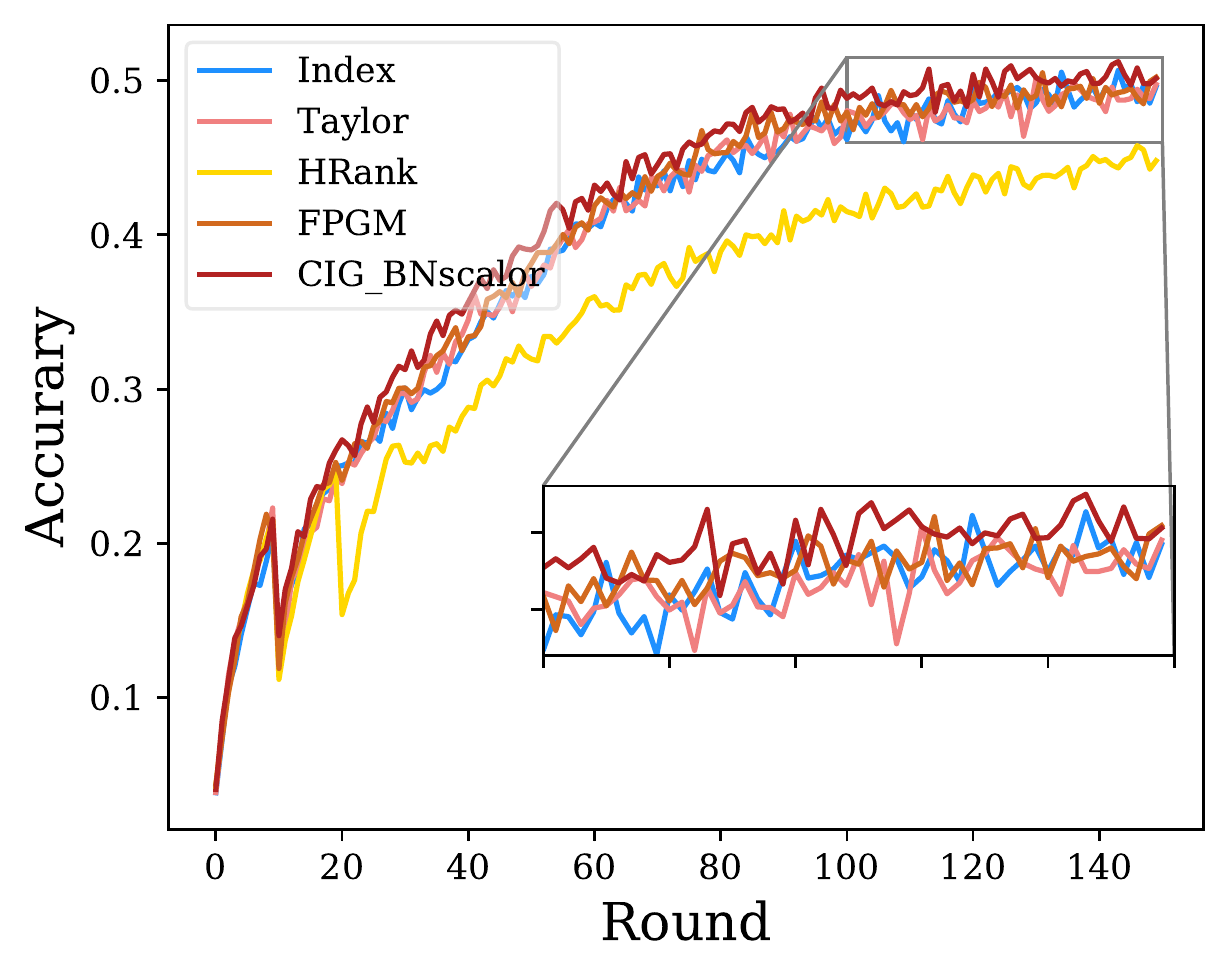}
    }

    \caption{The investigation of distributed pruning principles (CIFAR100). In (a) and (b), we show ablation experiments of \emph{Index} pruning method on IID dataset (a) and Non-IID ($s$=80) dataset (b). In (c), we show the remaining network similarity of different sub-models under multiple state-of-the-art pruning methods as pruning proceeds. In (d) and (e), we show results of multiple state-of-the-art pruning methods on the IID dataset (d) and Non-IID ($s$=80) dataset (e).}
    \label{figure:investigation of distributed pruning criteria}
\end{figure*}

We show results of the above three variants and \emph{Index} in Fig. \ref{figure:explore_index_methods_iid} and Fig. \ref{figure:explore_index_methods_iid8}. As we can see, both the experiments on the IID dataset and the Non-IID dataset showed the same results. \emph{No identical} has the worst result, and the global model does not converge, which reveals that \emph{identical} is the crucial reason. Followed by \emph{No constant}, the global model converges but converges to a lower accuracy, which reveals that \emph{constant} is also important. \emph{No adjacent} behaves almost the same as \emph{Index}, which reveals that \emph{adjacent} doesn't matter at all. 

To further confirm the importance of \emph{identical} and \emph{constant}, we investigate the remaining sub-model similarity of workers with the same pruned rate at each round. We define the similarity by the mean value of the ratio of the units intersection size to the units union size per layer, as in Eq. \ref{equation:similarity}. Several state-of-the-art model pruning methods are applied in AdaptCL to prune sub-model, including \emph{Taylor} \cite{molchanov2016pruning}, \emph{FPGM} \cite{he2019filter}, and \emph{HRank} \cite{lin2020hrank}. The results of the remaining network similarity are shown in Fig. \ref{figure:diff_method_saved_model}, and the global model accuracies in AdaptCL are shown in Fig. \ref{figure:diff_pruning_methods_iid} (IID) and  Fig. \ref{figure:diff_pruning_methods_iid8} (Non-IID). As we can see, the remaining sub-models of \emph{HRank} have the lowest similarity, i.e., the least \emph{identical} and \emph{constant} between workers, and its global models have the lowest accuracy regardless of the data distribution. The remaining sub-model similarity of \emph{FPGM} is slightly higher than that of \emph{Taylor}, and its accuracy is also slightly higher than that of \emph{Taylor}. So we conclude that \emph{identical} and \emph{constant} are crucial for distributed pruning. 

\begin{Definition}
\setlength{\abovedisplayskip}{0.6pt}
Given global index ${I}_{w}^{t}$ of worker $w$' sub-model at round t, the similarity of two workers' ($w1$, $w2$) remaining networks is defined as:
\begin{equation}
    \begin{aligned}
       \frac{1}{N} \sum_{n=1,2,...N} \frac{\vert {I}_{w1}^{t}[n] \cap {I}_{w2}^{t}[n] \vert  }{\vert{I}_{w1}^{t}[n] \cup {I}_{w2}^{t}[n] \vert}\\
    \end{aligned}
        \label{equation:similarity}
    \end{equation}
where ${I}_{w}^{t}[n]$ is the global index of the $n$th layer. $N$ is the number of layers. $\vert {I}_{w1}^{t}[n] \cup {I}_{w2}^{t}[n] \vert$ gets the size of the set.
\end{Definition}

Why are they crucial? \cite{sun2020learning} indicates that sharing the same model structure results in better performance when tasks are more similar in multi-task training. We believe that in collaborative learning, the tasks of individual workers are extremely similar, although workers may have different data distributions. \emph{      Identical} and \emph{constant} ensure maximum similarity between sub-models by guaranteeing that ${I}_{w1}^{t} \subset {I}_{w2}^{t}$, if ${\gamma}_{w1}^{t} < {\gamma}_{w2}^{t}$, i.e., the sub-model with more retention always covers the sub-model with less retention. 

Having discovered and understood the important principles for distributed pruning, we only need to keep the principles in mind when designing the pruning approach. Data-dependent importance evaluation methods can lead to disagreement between sub-models, so a data-independent method is required. Considering that different layers express different semantics, we design a \emph{constant}, \emph{identical}, and \emph{global} (\emph{CIG}) pruning approach. Batch normalization \cite{ioffe2015batch} has been widely used for fast convergence and generalization. Its scaling factor has been studied as a global importance evaluation method \cite{liu2017learning,ye2018rethinking} and is data-independent. So we adopt the BN scaling factors of the aggregated global model at the first pruning to measure the importance of all units and keep the importance order for all workers at each pruning. We prune units below a global importance threshold across all layers, which is defined from the pruning budget. Thus we get a \emph{CIG} pruning approach, named \emph{CIG-BNscalor}. As shown in Fig. \ref{figure:diff_pruning_methods_iid} and  Fig. \ref{figure:diff_pruning_methods_iid8}, \emph{CIG-BNscalor} is superior to the state-of-the-art methods.

 Besides, when to make the pruning is also a concern for pruning in the training process. We use iterative pruning, pruning after each pruning interval ($PI$), and pruning from the very beginning of training, making the overall training process as time-saving as possible and leaving more rounds to recover the model.

\section{Experiments}

\subsection{Experimental Settings}
{\bf Datasets, models and baselines}. We evaluate AdaptCL on CIFAR10, CIFAR100 \cite{krizhevsky2009learning} and Tiny-ImageNet\footnote{Tiny-ImageNet visual recognition challenge, https://tiny-imagenet.herokuapp.com.}. We train a
variation of VGG16 on CIFAR10 and CIFAR100, as in \cite{lin2020hrank}, and ResNet50 \cite{he2016deep} on Tiny-ImageNet. We compare AdaptCL with five methods, including FedAVG \cite{mcmahan2017communication}, FedAVG-S, FedAsync-S \cite{xie2019asynchronous}, SSP-S\cite{dai2015high,ho2013more} and DC-ASGD-a-S \cite{zheng2017asynchronous}. The suffix \emph{-S} represents the use of sparse training approach mentioned in Sec. \ref{section:model training}.

{\bf Non-IID setting}. We partition the Non-IID dataset in the same way as \cite{Karimireddy2020SCAFFOLDSC}. We divide the (1-$s$\%) of the IID dataset equally to each worker, and the remaining $s$\% of the IID dataset is sorted by the label and divided sequentially to each worker. So we make sure that each worker has the same amount of data but a different number for each class.

{\bf Heterogeneous setting}. The heterogeneity $H$ is defined based on the distribution of worker update time ${\phi}_{w}$ (including training time and communication time) as Eq. \ref{equation:heter}.

\begin{equation}
\begin{aligned}
    &H = 1-\frac{1}{W-1}\sum_{w=1}^{W-1} \frac{{\phi}_{W}}{{\phi}_{w}} \\
    &Assume \ {\phi}_{W} = Min({\phi}_{1},...,{\phi}_{W})
\end{aligned}
    \label{equation:heter}
\end{equation}

In our simulation experiments, since all workers are on the same device, training time of workers is not heterogeneous. So we set the bandwidth of workers differently to achieve needed heterogeneity.

{\bf Configurations}. We build AdaptCL using PyTorch, and experiment on NVIDIA V100. In our experiments, $W$ = 10, $PI$ = 10. Detail setting can be found in Appendix \ref{appendix:Experiment setting}.

\subsection{Results and Analysis}
In this section, we set ${\rho}_{max}$ = 0.5, ${\gamma}_{min}$ = 0.1, $B_{max}$ = 5MB, $\sigma$ = 2, and the corresponding heterogeneity is about 0.32. We show top-1 test accuracy (Acc) and total training time (Time) in tables. For FedAsync, SSP and DC-ASGD-a, we report the best accuracy of aggregations and the corresponding finished time for that round.

\begin{table}[htbp]
\begin{center}
\resizebox{!}{2.3cm}{
\begin{tabular}{cccccc}
 \toprule
 \multirow{2}{*}{Dataset} & \multirow{2}{*}{Framework} & \multicolumn{2}{c}{IID($s$=0)} &  \multicolumn{2}{c}{Non-IID($s$=80)}\\
 \cmidrule(r) {3-4} \cmidrule(r) {5-6}
 &  & Acc(\%) & Time(min) & Acc(\%) & Time(min) \\
\midrule
\multirow{4}{*}{CIFAR10} 
 &  FedAVG \cite{mcmahan2017communication} &84.91  &270.77   & 72.06 &271.28   \\
 &  FedAVG-S \cite{mcmahan2017communication} &87.18 &279.37  &79.60 &279.86  \\
 &  FedAsync-S \cite{xie2019asynchronous} &81.70 &264.72  &69.93 & 203.40 \\
 &  SSP-S \cite{ho2013more} &85.61 &273.08  &75.45 & 282.48 \\
 &  DC-ASGD-a-S \cite{zheng2017asynchronous} &79.10 & 638.09& 58.13 & 869.61 \\
 &  AdaptCL(Ours) & {\bf 87.35} & {\bf 171.80}  & {\bf 80.90} &  {\bf 157.58} \\
\cmidrule {2 -6}
\multirow{4}{*}{CIFAR100}
 &  FedAVG \cite{mcmahan2017communication} & 55.80 & 270.40 & 51.26 & 271.53  \\
 &  FedAVG-S \cite{mcmahan2017communication} &61.65 &280.67  & 51.67 & 280.63 \\
 &  FedAsync-S \cite{xie2019asynchronous} &50.67 & 211.22 &42.98 & 202.44 \\
 &  SSP-S \cite{ho2013more} &60.66 & 283.27  &50.60 & 285.03 \\
 &  DC-ASGD-a-S \cite{zheng2017asynchronous} &49.39 &849.06 &36.15 & 667.69 \\
 &  AdaptCL(Ours) & {\bf 62.17 } & {\bf 152.45} &{\bf 51.85} & {\bf 154.32} \\
\bottomrule
\end{tabular}}
\end{center}
\caption{Experimental results of VGG16 on CIFAR.}
\label{table:VGG}
\end{table}

{\bf CIFAR}. The results of VGG16 on CIFAR10 and CIFAR 100 are presented in Tab. \ref{table:VGG}. More results and analysis can be found in Appendix \ref{appendix:performance}. For FedAVG and FedAVG-S, sparse training results in some improvement in accuracy because it is equivalent to regularization but also introduces more computational overhead, which slows down the training process. FedAsync, SSP, and DC-ASGD all belong to the asynchronous way, and more or less gradient staleness causes the worker data not to be fully utilized, so the accuracy is lower than that of the synchronous way, i.e., FedAVG-S. And the impact of gradient staleness increases gradually with the difficulty of the task, the degree of data Non-IID. The increase in time for SSP comes from the fact that the server needs to do $W*T$ aggregations and that for DC-ASGD comes from the increase in communication time due to the small local training epochs $E$ set.

\begin{figure}[htbp]
    \centering
    \subfigure[Test accuracy vs. round]{
        \label{figure:accround}
        \includegraphics[width=.5\linewidth]{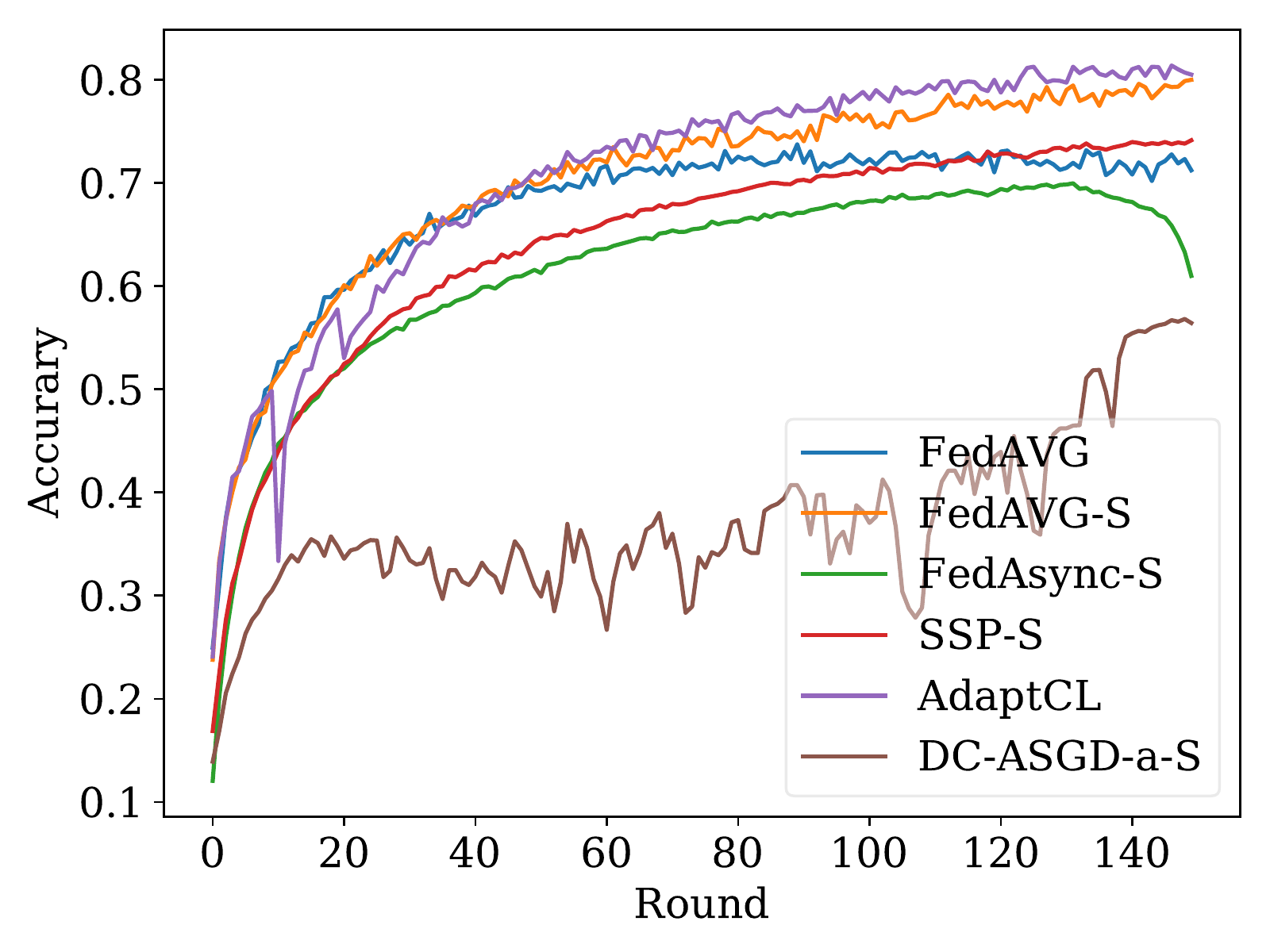}
    }
    \subfigure[Test accuracy vs. time]{
    \label{figure:timeacc}
	\includegraphics[width=.5\linewidth]{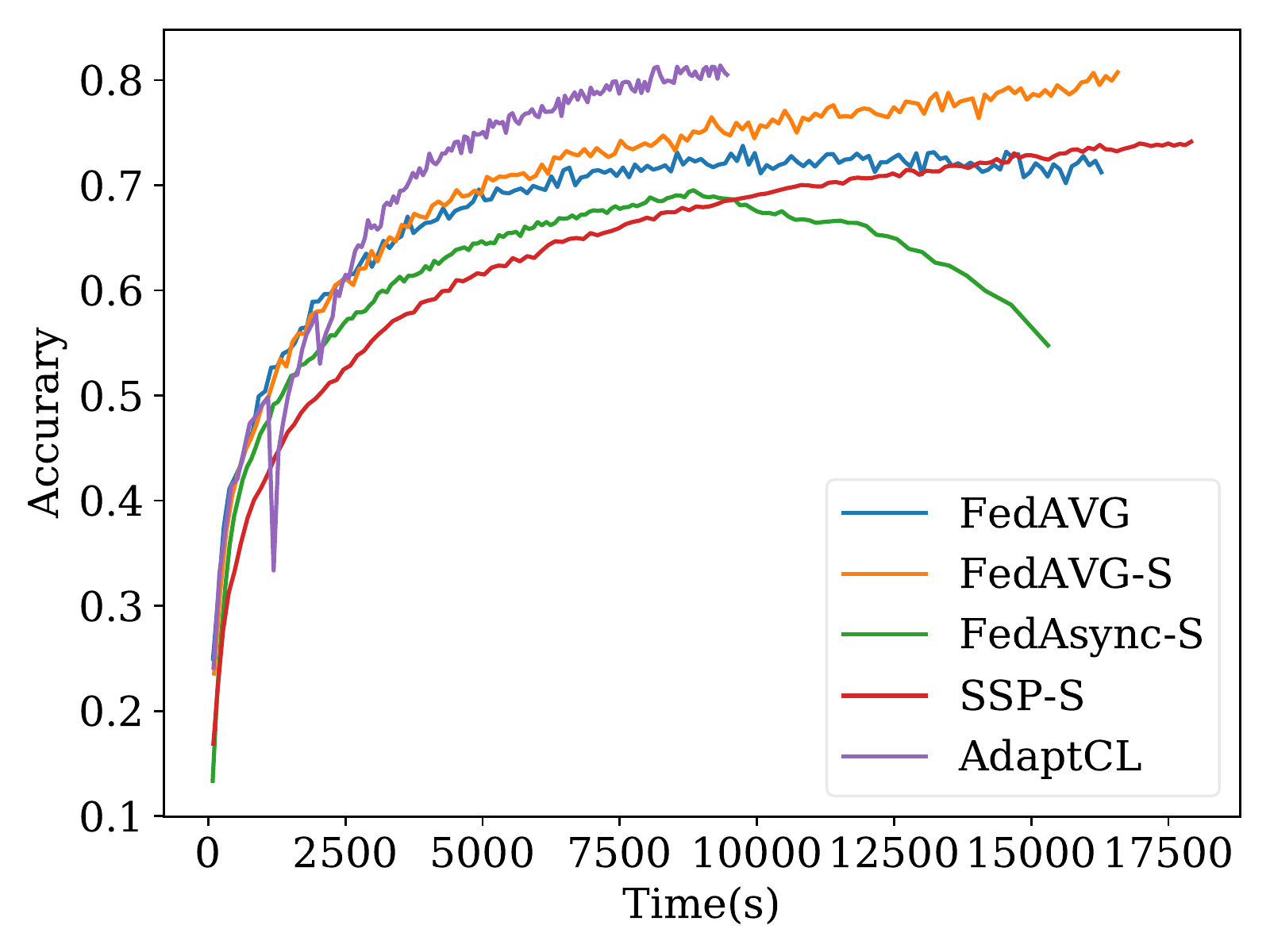}
    }
    \caption{Comparison with baselines (CIFAR10, Non-IID, $s$=80).}
    \label{figure:comparison}
\end{figure}

AdaptCL achieves the best performance on the four data datasets. First, AdaptCL exceeds FedAVG-S in accuracy by 0.17\%, 1.3\%, 0.52\% and 0.18\% for IID CIFAR10, Non-IID CIFAR10, IID CIFAR100, and Non-IID CIFAR100, respectively. As we can see from Fig. \ref{figure:accround}, for AdaptCL, there is a slight loss in accuracy after the initial pruning process, but it quickly recovers and catches up or even surpasses other methods. The improvement probably has to do with the fact that the pruned parameters are treated as zero in the aggregation, and most of them are optimized toward zero as well. So pruning accelerates the optimization of most of the pruned parameters. Second, AdaptCL has the shortest training time on all datasets due to the reduction of model parameters and FLOPs. Specifically, compared to  FedAVG-S, AdaptCL saves training time by 38.5\%, 43.7\%, 45.7\%, and 45.0\%, respectively. Taken together, AdaptCL achieves higher accuracy in the same amount of time, i.e., more efficient training (Fig. \ref{figure:timeacc}). 

\begin{table}[htbp]
\begin{center}
\resizebox{!}{1.3 cm}{
\begin{tabular}{cccccc}
 \toprule
 \multirow{2}{*}{Dataset} & \multirow{2}{*}{Framework} & \multicolumn{2}{c}{IID($s$=0)} &  \multicolumn{2}{c}{Non-IID($s$=80)}\\
  \cmidrule(r) {3-4} \cmidrule(r) {5-6}
 &  & Acc(\%) & Time(min) & Acc(\%) & Time(min) \\
\midrule
\multirow{4}{*}{\makecell[c]{Tiny \\ ImageNet}} 
 &  FedAVG \cite{mcmahan2017communication} &61.08  &1206.51  & 55.40 &1206.51  \\
 &  FedAVG-S \cite{mcmahan2017communication}&{\bf 61.54} &1252.86  &{\bf 55.56} &1251.35  \\
 &  FedAsync-S \cite{xie2019asynchronous} &42.42 &1314.28  &36.65 &1107.21\\
 &  SSP-S \cite{ho2013more} &59.09 & 1029.25  &54.11 & 1024.26 \\
 &  AdaptCL(Ours) & 58.26 & {\bf 783.94}  & 51.61 & {\bf 777.60} \\
\bottomrule
\end{tabular}}
\end{center}
\caption{Experimental results of ResNet50 on Tiny-ImageNet.}
\label{table:resnet}
\end{table}

{\bf Tiny-ImageNet}. The results of ResNet50 on Tiny-ImageNet are presented in Tab. \ref{table:resnet}. AdaptCL obtains 58.26\% accuracy on the IID dataset and 51.61\% accuracy on the Non-IID dataset with 37.43\% and 37.85\% training time savings compared with FedAVG-S. As the task's difficulty increased, AdaptCL still shows the high overall performance, which exhibits its robustness.

\subsection{Sensitivity Evaluation}

\begin{table}[htbp]
\begin{center}
\resizebox{!}{1.3 cm}{
\renewcommand\tabcolsep{5pt}
\begin{tabular}{ccccccc}
 \toprule
 \multirow{2}{*}{$H$($\sigma$)} & \multicolumn{3}{c}{Non-IID CIFAR10} &  \multicolumn{3}{c}{Non-IID CIFAR100}\\
  \cmidrule(r) {2-4} \cmidrule(r) {5-7}
 & $\Delta$Acc(\%) & Time & Param $\downarrow$ & $\Delta$Acc(\%) & Time & Param $\downarrow$\\
\midrule
 0.32(2) &+1.3  &1.78x  & 47.65\%  & +0.18 &1.81x &49.99\% \\
 0.62(5) &+0.32 &3.15x & 67.25\% & -0.68  &3.09x &67.98\%  \\
 0.76(10) &+0.92 &4.85x &76.62\% & -1.11 &4.80x &76.54\%\\
 0.87(20) &-0.04 & 6.20x & 81.46\%  & -0.72 & 6.19x & 82.18\% \\
\bottomrule
\end{tabular}}
\end{center}
\caption{Performance of AdaptCL on CIFAR10 and CIFAR100 comparing to FedAVG-S under different heterogeneity.}
\label{table:differnet_hetero}
\end{table}

{\bf Performance under different heterogeneity}. By setting a different ratio of $\sigma$, we obtained different heterogeneity. The performance of AdaptCL on CIFAR10 and CIFAR100 (Non-IID, $s$=80) is reported in Tab. \ref{table:differnet_hetero}, taking  FedAVG-S as comparison. As heterogeneity increases, fewer parameters are left behind, and the accuracy decreases gradually. When at a high heterogeneity $H$=0.87, i.e., the longest update time is 20 times the shortest one, AdaptCL achieves a 6.2x acceleration of training, with only a small loss in accuracy of 0.04\% and 0.72\%, respectively. 

\begin{figure}[htbp]
    \centering
    \subfigure[Maximum pruned rate ${\rho}_{max}$]{
        \label{figure:max_pruned}
        \includegraphics[width=0.5\linewidth]{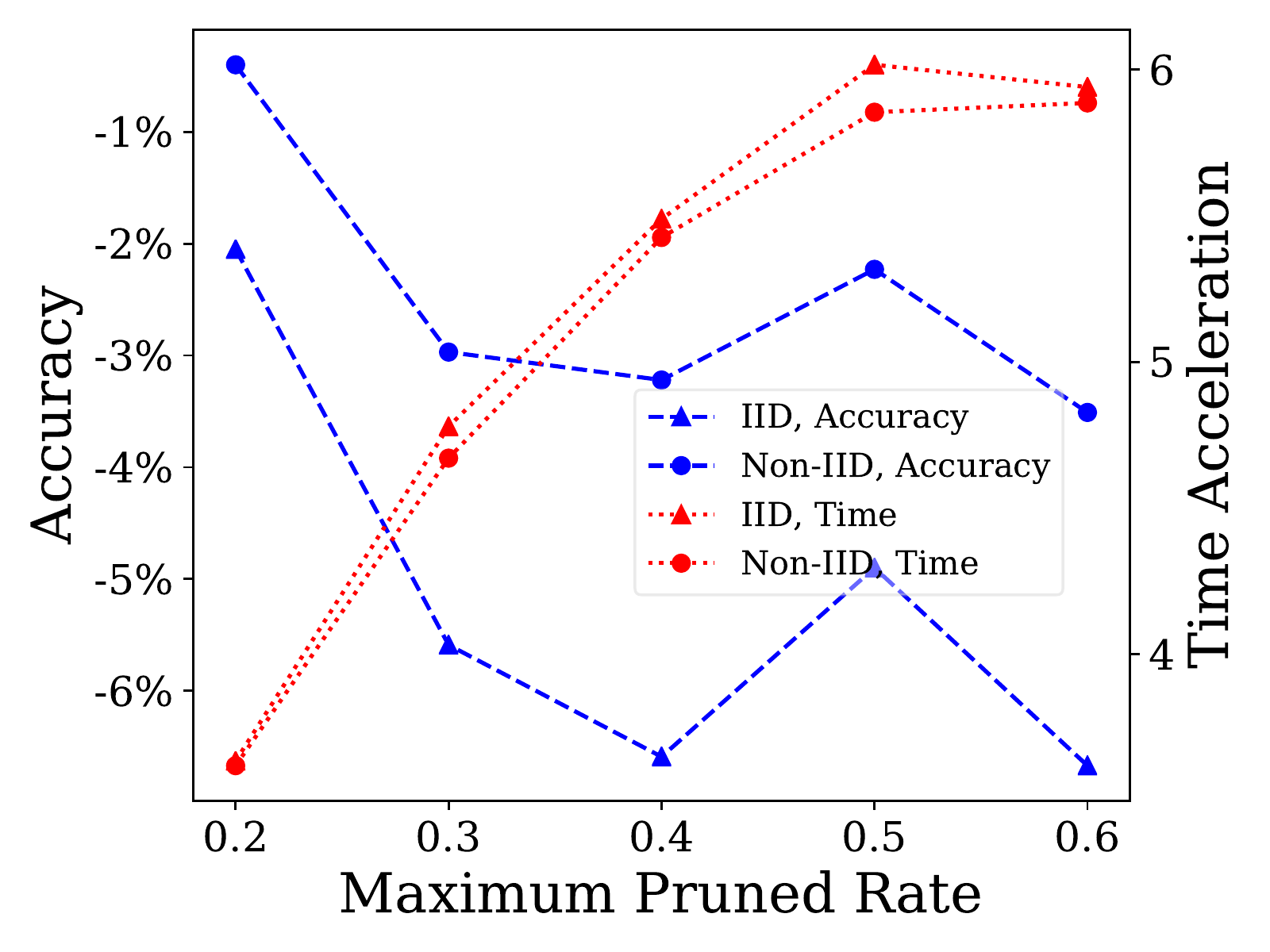}
    }
    \subfigure[Minimum retention ratio ${\gamma}_{min}$]{
    \label{figure:min_saved}
	\includegraphics[width=.5\linewidth]{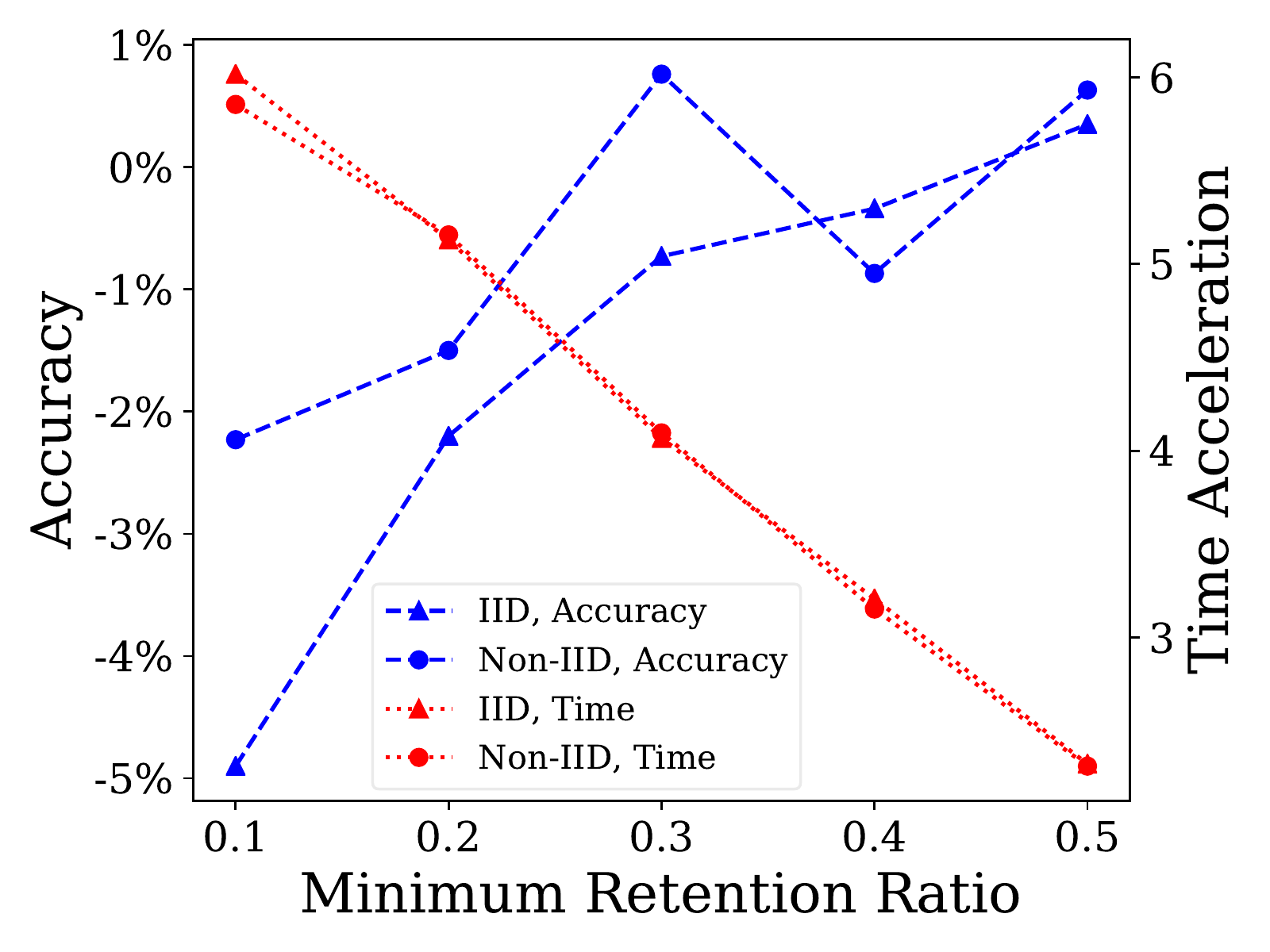}
    }
    \caption{Performance of AdaptCL on CIFAR100 ($H$=0.87, $B_{max}$=30) comparing to FedAVG-S under different controlling parameters.}
    \label{figure:controlling parameters}
\end{figure}

{\bf Impact of controlling parameters}.
Here we analyze the effect of the controlling parameters inside AdaptCL on performance.  Fig. \ref{figure:controlling parameters} reports the performance of accuracy and time acceleration on CIFAR100 under different controlling parameters, taking FedAVG-S as a comparison.

For maximum pruned rate ${\rho}_{max}$ (Fig. \ref{figure:max_pruned}), the model achieves better accuracy when there are fewer cuts per pruning, but more rounds are required for the pruning, causing the overall time to rise  (e.g., -0.4\% with 3.62x when ${\rho}_{max}$ = 0.2 vs. -2.23\% with 5.85x when ${\rho}_{max}$ = 0.5, Non-IID). For minimum retention ratio ${\gamma}_{min}$ (Fig. \ref{figure:min_saved}), the higher the ratio, the more accurate the model is, but more parameters are left behind resulting in higher overall time (e.g., -4.9\% with 6.01x when ${\gamma}_{min}$ = 0.1 vs. 0.35\% with 2.32x when ${\gamma}_{min}$ = 0.5, IID). This indicates that the size of parameters left behind is closely related to the accuracy of the model. 

As we can see from the results above, the controlling parameters allow us to do a trade-off between accuracy and time. When accuracy is more of a concern, a low maximum pruned rate as well as a high minimum retention ratio can be set, and vice versa.

\begin{figure}[htbp]
    \centering
    \subfigure[CIFAR10, IID]{
        \label{figure:diff_iid}
        \includegraphics[width=0.5\linewidth]{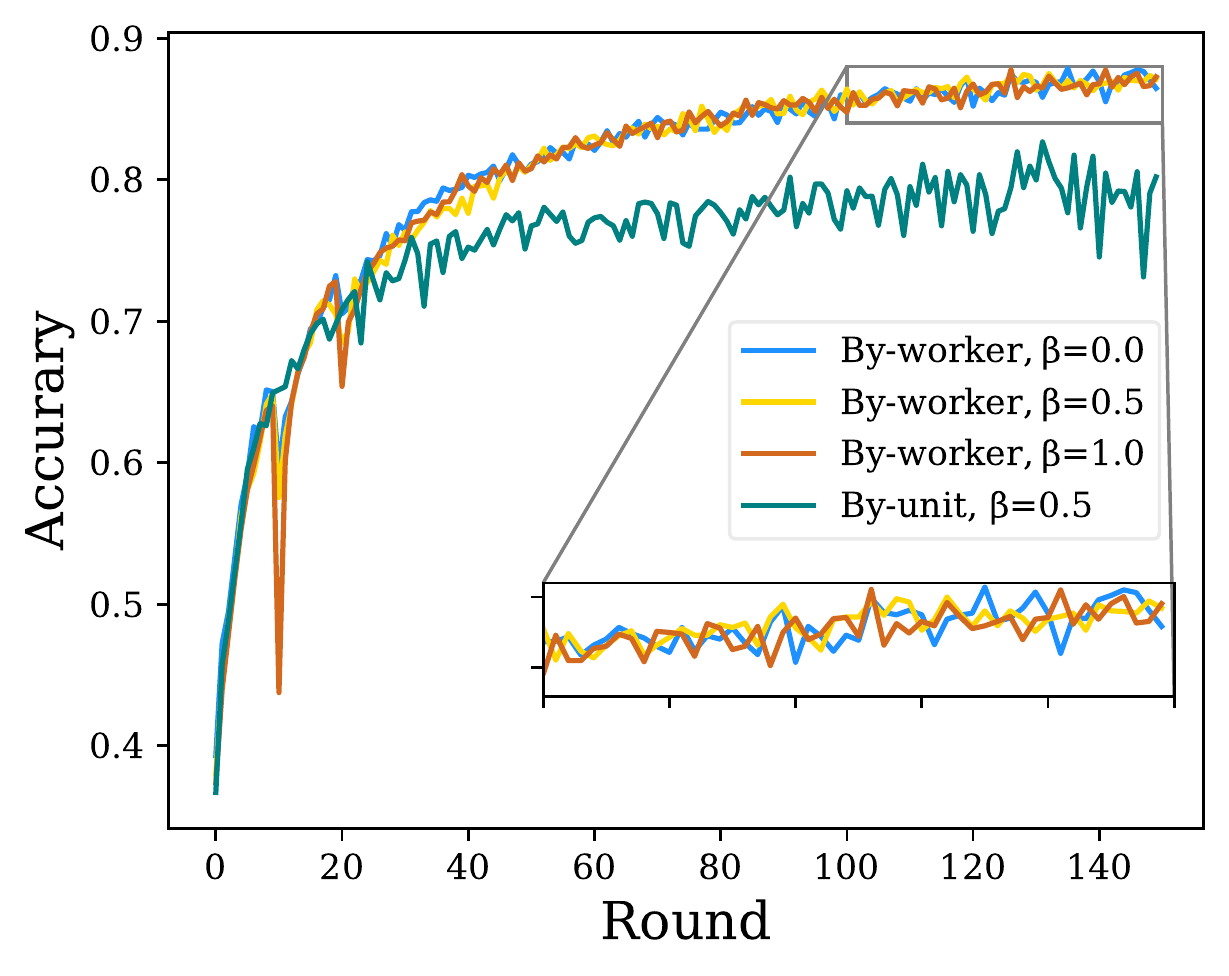}
    }
    \subfigure[CIFAR10, Non-IID, $s$=80]{
    \label{figure:diff_noniid}
	\includegraphics[width=.5\linewidth]{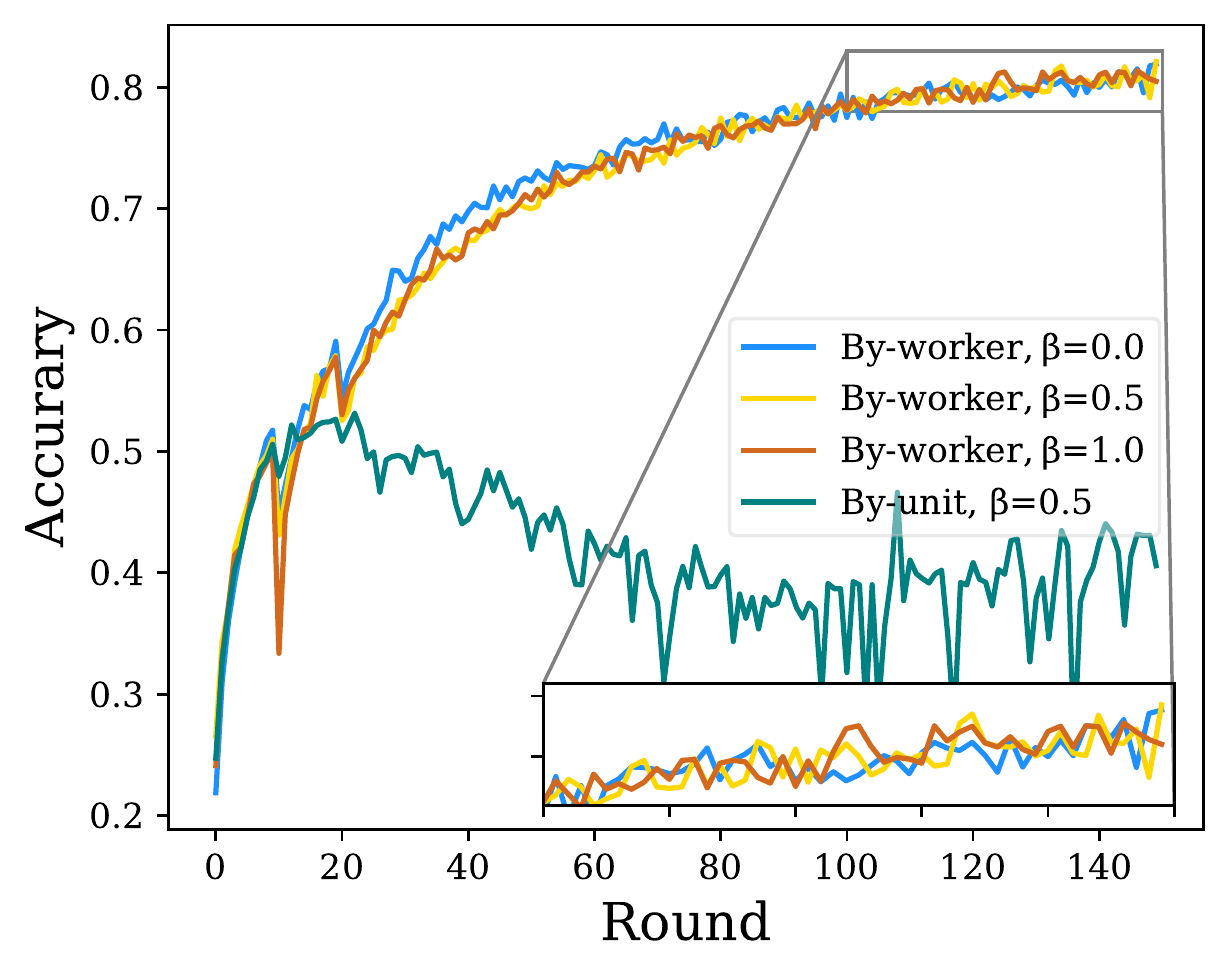}
    }
    \caption{Comparison of different pruning positions and model aggregation methods.}
    \label{figure:diff_round_acc}
\end{figure}

{\bf Impact of pruning position and model aggregating}. We report the performance under different pruning positions and model aggregation methods in Fig. \ref{figure:diff_round_acc}. For a fair comparison, we set the pruned rate per worker per round before the experiment (Appendix \ref{appendix:Experiment setting}). Overall, the pruning position has little effect on accuracy. When $\beta = 1.0$, since no local fine-tune is done after local pruning, accuracy dropped sharply at first but recovered quickly later. More analysis is deferred to the Appendix \ref{appendix:Internal performance}. When By-unit is used for model aggregation, the model does not have a drop in accuracy after pruning. Under IID dataset (Fig. \ref{figure:diff_iid}), accuracy continues to rise after pruning but soon stops rising. Worse, accuracy no longer rises after pruning under Non-IID dataset (Fig. \ref{figure:diff_noniid}). In our opinion, By-unit treats the pruned weights (those weights are pruned due to low values) as the mean of the unpruned weights of the other sub-models at the corresponding location. Thus the global model no longer reflects the information from local models.

\section{Discussion}
In this section, we discuss the limitations and potential future
directions of AdaptCL.

 First, due to the parallel optimization of the hardware and the small size of data used in the experiment, the training time is less sensitive to the pruning, resulting in the need to prune more parameters to reduce the communication time, which makes it more challenging to resolve this level of heterogeneity. When the worker's training time is sensitive to pruning, AdaptCL can resolve the heterogeneity by pruning fewer parameters and therefore achieves higher accuracy (the results running on CPU in Appendix \ref{appendix:Further Enhancements}). 
 Second, AdaptCL is entirely orthogonal to other acceleration methods, such as gradient quantization, and can be combined to achieve further acceleration (the results of AdaptCL+DGC \cite{lin2018deep} in Appendix \ref{appendix:Further Enhancements}).
Finally, with very high heterogeneity, AdaptCL cannot reduce the update time of all workers to the same level as the fastest worker. However, we can significantly reduce the training time down to a relatively low level by adjusting the minimum model retention ratio. 
 
\section{Conclusions}
In this paper, we propose a novel and efficient collaborative learning framework named AdaptCL, which generates an adaptive sparse sub-model dynamically from the global base model for each data holder based on its capability. By equipping capability-different workers with adaptive size models, all workers commit model updates near-synchronously, thus avoiding the dragger and staleness issues. We discuss in detail model training, pruning, and aggregation in the framework and the design of dynamic pruned rate learning algorithms that do not require prior capability-related information. Extensive experiments on various models and datasets demonstrate the efficiency of AdaptCL. In the future, we will do more theoretical research on distributed adaptive pruning and explore more efficient and precise pruning methods adapting to the dynamic heterogeneous environment.

{\small
\bibliographystyle{my_bib_style}
\bibliography{main}
}

\clearpage

\appendix
\begin{appendices}
\section*{Appendices}
The appendix is divided into five parts. Appendix \ref{appendix:AdaptCL details} is an additional description of AdaptCL. Appendix \ref{appendix:Experiment setting} provides a more detailed description of the experimental setting. Appendix \ref{appendix:performance} shows more comprehensive experimental results. Appendix \ref{appendix:Internal performance} is a more in-depth exploration of AdaptCL. Appendix \ref{appendix:Further Enhancements} demonstrates the potential and future directions of AdaptCL.

\section{AdaptCL Details}
\label{appendix:AdaptCL details}
\begin{figure}[htbp]
\begin{center}
  \includegraphics[width=0.9\linewidth]{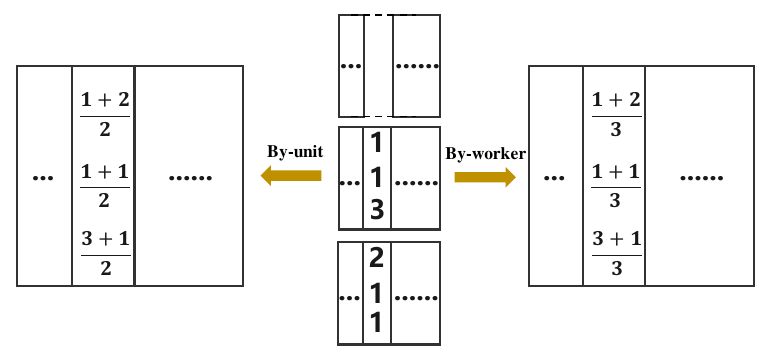}
\end{center}
  \caption{Two approaches to aggregate parameters. The three squares in the middle represent one weight matrix of three workers, respectively. The squares on either side represent the results of the two ways of aggregation. The top worker cuts out a unit, resulting in a missing column of the matrix. Here, W=3, $w^{'}$=2.}
\label{figure:modelagg}
\end{figure}

In this section we elaborate on AdaptCL in more detail.
{\bf Model aggregating}. By-worker and By-unit are illustrated in Fig. \ref{figure:modelagg}. 

{\bf Pruned Rate Learning}. Actually, the update time observed by the server is affected by random factors, such as fluctuations in bandwidth, fluctuations in training time, etc. As we discussed in Sec. \ref{section:network prune}, there is a pruning interval between our adjacent pruning, and the update time of all rounds in the interval will be averaged as the corresponding update time of the pruned model, which avoids the influence of some random factors and makes our system more stable.

The Newton interpolation may have the Runge phenomenon at higher orders causing larger errors. However, since approximately identical update time can be achieved with three or four rounds of pruning as shown in Fig. \ref{figure:roundtime} and Fig. \ref{figure:roundhetero}, the $n$ is small in our scenario, so the Runge phenomenon does not occur. According to Theorem of Interpolation Error \cite{sauer2006numerical}, the interpolation error is shown below. The ${f}_{w}^{-1}$ represents the actual function we want to approximate.

\begin{equation}
\setlength{\abovedisplayskip}{0.2pt}
\setlength{\belowdisplayskip}{0.2pt}
\begin{aligned}
    &{f}_{w}^{-1}({\phi}_{min})- \widetilde {f}_{w}^{-1}({\phi}_{min}) \\
    &= \frac{({\phi}_{min}-{\phi}_{w}^{0})...({\phi}_{min}-{\phi}_{w}^{n})}{n!}{{f}_{w}^{-1}}^{(n)}(c) \\
    & where \ min({\phi}_{w}^{0},...,{\phi}_{w}^{n})<c<max({\phi}_{w}^{0},...,{\phi}_{w}^{n})
\end{aligned}
\label{equation:prunedrate—error}
\end{equation}

\begin{figure}[htbp]
\begin{center}
  \includegraphics[width=0.9\linewidth]{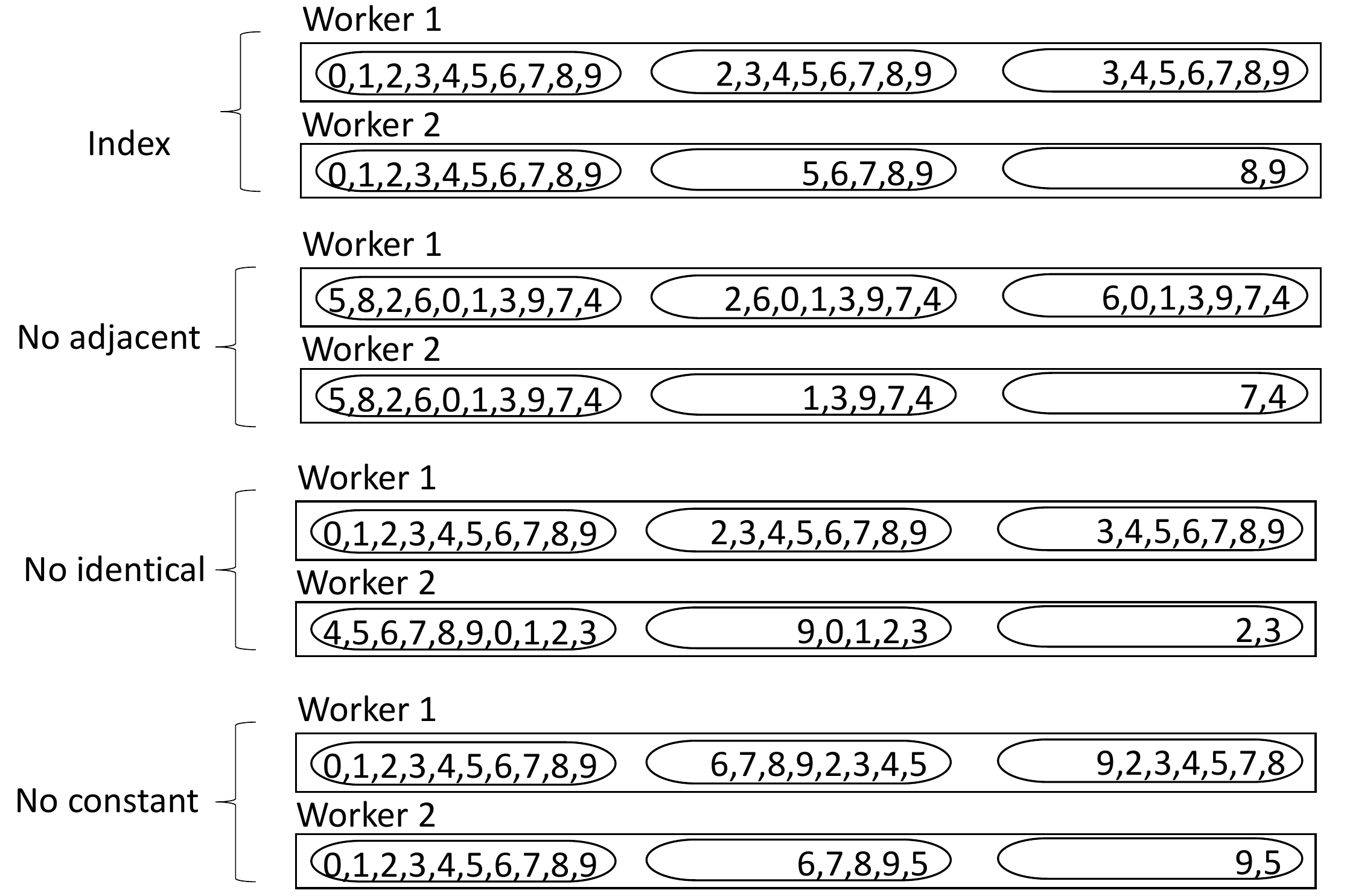}
\end{center}
  \caption{An example of \emph{Index} and three variants.}
\label{figure:pruning variant}
\end{figure}

{\bf Model Pruning}. We give an example of \emph{Index} and three variants in Fig. \ref{figure:pruning variant}. The sequence of numbers represents the determined order of unit pruning. The \emph{Index} method prunes in index order, with each worker following the same index order of the beginning at each round. \emph{No adjacent} generates the same random order for all workers. \emph{No identical} makes the initial order different between each node. \emph{No constant} makes the order of all workers per round start from a new but same position. 

\section{Experiment setting}
\label{appendix:Experiment setting}
In this section, we give specific settings in the experiment.

{\bf Datasets and models}. CIFAR10 and CIFAR100 consist of 50,000 training images and 10,000
validation images in 10 and 100 classes, respectively. Tiny-ImageNet has 200 classes, and each class contains 500 training images, 50 validation images.

{\bf Baselines}. FedAsync-S is an asynchronous FedAVG using the sparse training approach, where the aggregation weights for the local and global models are set in a polynomial way, and the hyperparameter $a$ is set to 0.5. The aggregation coefficient in SSP is set to $1/W$. The threshold $s$ in SSP is set to 2, 4, and 8, respectively, and we chose the best result as the result of SSP.
For FedAsync and SSP, each worker runs T rounds, resulting in $W*T$ aggregations, and we report the best accuracy of $W*T$ aggregations and the corresponding finished time for that round.

DC-ASGD is an asynchronous way of committing gradients, where the model is updated on the server-side. It compensates for the stale gradient to reduce its impact on the global model. After our grid search for parameters (\ref{table:DC-ASGD}), we found that the smaller the number of local update epoch $E$, the higher the accuracy, and the other parameters play a minimal role. This is because the smaller the $E$, the slower workers can get the latest model in a shorter interval, and the gradient staleness issue can be relatively weakened. However, this greatly increases the overall training time. So committing gradients is not efficient in collaborative learning, which is consistent with the conclusion in \cite{mcmahan2017communication}. We set ${\lambda}_{0}$=2.0, ${m}$=0.95, $E$=0.5 ,$\eta$=0.01 for it in experiments.

\begin{table}[htbp]
\begin{center}
\resizebox{!}{1.8 cm}{
\begin{tabular}{ccc}
\toprule
DC-ASGD-a & Acc & Time(min) \\
\midrule
${\lambda}_{0}$=2.0, ${m}$=0.95, $E$=2 ,$\eta$=0.01  & 67.64\% & 261.07\\
${\lambda}_{0}$=20.0, ${m}$=0.95, $E$=2 ,$\eta$=0.01 & 67.15\% &  262.05\\
${\lambda}_{0}$=100.0, ${m}$=0.95, $E$=2 ,$\eta$=0.01 & 67.17\% & 261.88\\
${\lambda}_{0}$=1.0, ${m}$=0.95, $E$=2 ,$\eta$=0.01 & 66.56\% & 262.03 \\
${\lambda}_{0}$=2.0, ${m}$=0.0, $E$=2 ,$\eta$=0.01 & 66.99\% & 261.52 \\
${\lambda}_{0}$=2.0, ${m}$=0.95, $E$=1 ,$\eta$=0.01 & 74.08\% & 521.74 \\
${\lambda}_{0}$=2.0, ${m}$=0.95, $E$=1 ,$\eta$=0.1 & 75.20\% & 511.64 \\
${\lambda}_{0}$=2.0, ${m}$=0.95, $E$=1 ,$\eta$=0.001 & 49.65\% & 486.47 \\
${\lambda}_{0}$=2.0, ${m}$=0.95, $E$=0.5 ,$\eta$=0.01 & 79.10\% & 638.09 \\
\bottomrule
\end{tabular}}
\end{center}
\caption{Performance of DC-ASGD-a on IID CIFAR10 under different parameters.}
\label{table:DC-ASGD}
\end{table}

{\bf Heterogeneous setting}. 
A measure of the degree of heterogeneity in a collaborative system consists of three components: the ratio of the longest update time to the shortest update time $\sigma$, the distribution of update time of all workers, and the ratio of training time and communication time per worker, where the first two are knowable for the server. From the server's point of view, the indicator of heterogeneity $H$ is defined as Eq. \ref{equation:heter}.

The following describes the setting of the initial heterogeneity of the simulation. In our simulation experiments, since all workers are on the same device, the training time $t_{train}$ of workers is not heterogeneous. So we set the bandwidth of workers differently to achieve needed heterogeneity. 

First, we obtain $t_{train}$ by measuring the sparse training time per round and calculate model size $s_{model}$. Second, we set the bandwidth of the fastest worker $B_{max}$ and the ratio $\sigma$ so that we get the update time of the slowest and fastest workers. The update time of the remaining workers is uniformly distributed between them, as shown in Eq. \ref{equation:update_time}. Here, worker $W$ is the fastest worker. Finally, we can calculate the bandwidth set for each worker by Eq. \ref{equation:bandwidth}. The corresponding heterogeneity is as shown in Eq. \ref{equation:corrHeter}. We use the same set of bandwidth settings for all methods of comparison.

\begin{equation}
\begin{aligned}
{\phi}_{w}  &=  (\frac{2*s_{model}}{B_{max}} + t_{train}) *(1+\frac{\sigma -1 }{W-1}(W-w)) \\
\end{aligned}
\label{equation:update_time}
\end{equation}

\begin{equation}
\begin{aligned}
     B_{w} = \frac{2*s_{model}}{{\phi}_{w}-t_{train}}
\end{aligned}
\label{equation:bandwidth}
\end{equation}

\begin{equation}
\begin{aligned}
     H = 1-\frac{1}{W-1}\sum_{w=1}^{W-1}\frac{1}{1+\frac{\sigma -1 }{W-1}(W-w)}
\end{aligned}
\label{equation:corrHeter}
\end{equation}

Next, we show the bandwidth we set. For VGG16 on CIFAR10 and CIFAR100, we set bandwidth of workers as in Tab. \ref{table:5Mheter} and Tab. \ref{table:30Mheter}. For ResNet50 on Tiny-ImageNet, we set the bandwidth of workers as in Tab. \ref{table:5MResnetheter}.

\begin{table}[htbp]
\begin{center}
\resizebox{!}{1.2 cm}{
\begin{tabular}{ll}
\toprule
$H$($\sigma$) & Bandwidth (MB) \\
\midrule
0.32(2) & 1.63, 1.77, 1.93, 2.11, 2.34, 2.62, 2.97, 3.43, 4.07, 5\\
0.62(5) & 0.55, 0.60, 0.68, 0.77, 0.90, 1.07, 1.34, 1.77, 2.62, 5 \\
0.76(10) & 0.25, 0.28, 0.32, 0.37, 0.44, 0.54, 0.70, 0.98, 1.64, 5 \\
0.87(20) & 0.12, 0.14, 0.16, 0.18, 0.22, 0.27, 0.35, 0.51, 0.94, 5  \\
\bottomrule
\end{tabular}}
\end{center}
\caption{Bandwidth setting of workers when $B_{max}$=5M (VGG16).}
\label{table:5Mheter}
\end{table}

\begin{table}[htbp]
\begin{center}
\resizebox{!}{1.2 cm}{
\begin{tabular}{ll}
\toprule
$H$($\sigma$) & Bandwidth (MB) \\
\midrule
0.32(2) & 3.6, 4.0, 4.5, 5.14, 6.0, 7.1, 8.78, 11.48, 16.61, 30\\
0.62(5) & 1.0, 1.12, 1.27, 1.47, 1.75, 2.15, 2.8, 4.0, 7.1, 30 \\
0.76(10) & 0.45, 0.50, 0.57, 0.67, 0.8, 1.0, 1.3, 2.0, 3.6, 30 \\
0.87(20) & 0.21, 0.24, 0.27, 0.32, 0.38, 0.48, 0.63, 0.94, 1.83, 30  \\
\bottomrule
\end{tabular}}
\end{center}
\caption{Bandwidth setting of workers when $B_{max}$=30M (VGG16).}
\label{table:30Mheter}
\end{table}

\begin{table}[htbp]
\begin{center}
\resizebox{!}{0.55 cm}{
\begin{tabular}{ll}
\toprule
$H$($\sigma$) & Bandwidth (MB) \\
\midrule
0.32(2) & 0.997, 1.095, 1.213, 1.360, 1.547, 1.796, 2.138, 2.642, 3.458, 5.0\\
\bottomrule
\end{tabular}}
\end{center}
\caption{Bandwidth setting of workers when $B_{max}$=5M (ResNet50).}
\label{table:5MResnetheter}
\end{table}

{\bf Configurations} We build AdaptCL using PyTorch, and experiment on NVIDIA V100. In our experiments, $W$ = 10, $PI$ = 10, $\alpha$ = 2, $\beta$ = 1.0. For CIFAR10 and CIFAR100, $T$ = 150, $E$ = 2. For Tiny-ImageNet, $T$ = 250, $E$ = 1. We use a learning rate of 0.01 and 0.1 for CIFAR and ImageNet, and a mini-batch size of 64 for both. The weight
decay in sparse training is set to 5e-4. We adopt the same method in \cite{lym2019prunetrain} to set sparse coefficient $\lambda$ by sparsification strength. Sparsification strength is set to 0.9 for CIFAR10, IID CIFAR100 and 0.1 for the others. For VGG16, we do not prune the last fully connected layer. For ResNet50, we do not prune the first convolutional layer and the last layer of each residual block. 

\begin{table}[htbp]
\begin{center}
\resizebox{!}{1.4 cm}{
\begin{tabular}{ll}
\toprule
round & pruned rate \\
\midrule
10 & 0.5, 0.3, 0.2, 0.3, 0.3, 0.2, 0.3, 0.2, 0.2, 0.0\\
20 & 0.3, 0.2, 0.2, 0.2, 0.3, 0.3, 0.2, 0.2, 0.2, 0.0 \\
30 & 0.2, 0.1, 0.1, 0.1, 0.2, 0.2, 0.1, 0.0, 0.1, 0.0 \\
40 & 0.1, 0.0, 0.0, 0.0, 0.1, 0.0, 0.1, 0.0, 0.0, 0.0  \\
\bottomrule
\end{tabular}}
\end{center}
\caption{Pruned rate of workers.}
\label{table:pruned rate}
\end{table}

{\bf Pruned rate setting}. Due to fluctuations in training time, the pruned rate learning algorithm may give different pruned rates. Even with the same bandwidth settings, the final retention ratio of each worker's model is not precisely the same. For a fair comparison of different pruning positions, pruning methods, and model aggregation methods, we set the pruned rate per worker per round before the experiment, as shown in Tab. \ref{table:pruned rate}.

\begin{table*}[htbp]
\begin{center}
\resizebox{!}{1.5 cm}{
\renewcommand\tabcolsep{5pt}
\begin{tabular}{ccccccccc}
 \toprule
 \multirow{2}{*}{$H$($\sigma$)} & \multicolumn{4}{c}{$B_{max}$=5} &  \multicolumn{4}{c}{$B_{max}$=30}\\
  \cmidrule(r) {2-5} \cmidrule(r) {6-9}
 & $\Delta$Acc(\%) & Time & Param ($\downarrow$) & FLOPs ($\downarrow$) & $\Delta$Acc(\%) & Time & Param ($\downarrow$) & FLOPs ($\downarrow$)\\
\midrule
 0.32(2) &+0.17  &1.63x  & 29.94M(47.64\%) & 287.74M(54.23\%) & +0.01 &1.67x &27.71M(51.55\%) &267.94M(57.37\%) \\
 0.62(5) &-0.03 &3.07x &18.56M (67.56\%) &191.83M(69.48\%) & -0.2  &2.92x &14.00M(75.51\%) &170.34M(72.90\%) \\
 0.76(10) &-0.4 &4.62x &12.80M(77.62\%) &157.97M(74.87\%) & -0.53 &4.54x &10.13M(82.29\%)
 &142.73M(77.29\%) \\
 0.87(20) &-0.5 & 6.19x & 10.08M(82.37\%) &140.11M(77.71\%)  & -1.38 &6.06x &8.16M(85.73\%)
 &124.11M(80.26\%) \\
\bottomrule
\end{tabular}}
\end{center}
\caption{Performance of AdaptCL on CIFAR10 (IID) comparing to FedAVG-S under different heterogeneity.}
\label{table:differnet_hetero_cifar10_iid}
\end{table*}

\begin{table*}[htbp]
\begin{center}
\resizebox{!}{1.5 cm}{
\renewcommand\tabcolsep{5pt}
\begin{tabular}{ccccccccc}
 \toprule
 \multirow{2}{*}{$H$($\sigma$)} & \multicolumn{4}{c}{$B_{max}$=5} &  \multicolumn{4}{c}{$B_{max}$=30}\\
  \cmidrule(r) {2-5} \cmidrule(r) {6-9}
 & $\Delta$Acc(\%) & Time & Param ($\downarrow$) & FLOPs ($\downarrow$) & $\Delta$Acc(\%) & Time & Param ($\downarrow$) & FLOPs ($\downarrow$)\\
\midrule
 0.32(2) &+1.3  &1.78x  & 29.94M(47.65\%) & 262.78M(58.20\%) & +0.56 &1.35x &36.47M(36.23\%) &340.41M(45.85\%) \\
 0.62(5) &+0.32 &3.15x &18.73M (67.25\%) &198.80M(68.37\%) & +0.4  &3.10x &12.99M(77.29\%) &159.70M(74.59\%) \\
 0.76(10) &+0.92 &4.85x &13.37M(76.62\%) &160.44M(74.48\%) & +0.01 &4.59x &9.93M(82.63\%)
 &141.72M(77.46\%) \\
 0.87(20) &-0.04 & 6.20x & 10.60M(81.46\%) &143.97M(77.24\%)  & -0.56 & 6.33x &8.04M (85.95\%)
 &120.38M(80.85\%) \\
\bottomrule
\end{tabular}}
\end{center}
\caption{Performance of AdaptCL on CIFAR10 (Non-IID, $s$=80) comparing to FedAVG-S under different heterogeneity.}
\label{table:differnet_hetero_cifar10_iid8}
\end{table*}

\begin{table*}[htbp]
\begin{center}
\resizebox{!}{1.5 cm}{
\renewcommand\tabcolsep{5pt}
\begin{tabular}{ccccccccc}
 \toprule
 \multirow{2}{*}{$H$($\sigma$)} & \multicolumn{4}{c}{$B_{max}$=5} &  \multicolumn{4}{c}{$B_{max}$=30}\\
   \cmidrule(r) {2-5} \cmidrule(r) {6-9}
 & $\Delta$Acc(\%) & Time & Param ($\downarrow$) & FLOPs ($\downarrow$) & $\Delta$Acc(\%) & Time & Param ($\downarrow$) & FLOPs ($\downarrow$)\\
\midrule
 0.32(2) &+0.52  &1.63x  & 25.28M(55.93\%) & 248.26M(60.51\%) & -0.09 &1.50x &32.37M(43.57\%) &315.60M(49.80\%) \\
 0.62(5) &-0.88 &3.09x &17.48M (69.53\%) &186.73M(70.30\%) & -2.31  &3.16x &12.18M(78.76\%) &150.97M(75.99\%) \\
 0.76(10) &-1.78 &4.62x &13.08M(77.20\%) &160.13M(74.53\%) & -2.62 &4.56x &10.43M(81.82\%)
 &136.27M(78.33\%) \\
 0.87(20) &-3.79 & 6.17x & 10.35M(81.96\%) &136.36M(78.31\%)  & -4.9 &6.02x &8.57M(85.07\%)
 &121.74M(80.64\%) \\
\bottomrule
\end{tabular}}
\end{center}
\caption{Performance of AdaptCL on CIFAR100 (IID) comparing to FedAVG-S under different heterogeneity.}
\label{table:differnet_hetero_cifar100_iid}
\end{table*}

\begin{table*}[htbp]
\begin{center}
\resizebox{!}{1.5 cm}{
\renewcommand\tabcolsep{5pt}
\begin{tabular}{ccccccccc}
 \toprule
 \multirow{2}{*}{$H$($\sigma$)} & \multicolumn{4}{c}{$B_{max}$=5} &  \multicolumn{4}{c}{$B_{max}$=30}\\
   \cmidrule(r) {2-5} \cmidrule(r) {6-9}
 & $\Delta$Acc(\%) & Time & Param ($\downarrow$) & FLOPs ($\downarrow$) & $\Delta$Acc(\%) & Time & Param ($\downarrow$) & FLOPs ($\downarrow$)\\
\midrule
 0.32(2) &+0.18  &1.81x  & 28.69M(50.00\%) & 262.89M(58.18\%) & +0.11 &1.72x &18.11M(68.43\%) &193.70M(69.19\%) \\
 0.62(5) &-0.68 &3.09x &18.37M (67.98\%) &191.87M(69.48\%) & -0.13  &3.03x &12.37M(78.44\%) &151.68M(75.87\%) \\
 0.76(10) &-1.11 &4.80x &13.46M(76.54\%) &162.63M(74.13\%) & -2.57 &4.55x &9.75M(83.00\%)
 &132.64M(78.90\%) \\
 0.87(20) &-0.72 & 6.19x & 10.22M(82.18\%) &137.18M(78.18\%)  & -2.23 & 5.86x &8.93M (84.44\%)
 &125.12M(80.10\%) \\
\bottomrule
\end{tabular}}
\end{center}
\caption{Performance of AdaptCL on CIFAR100 (Non-IID, $s$=80) comparing to FedAVG-S under different heterogeneity.}
\label{table:differnet_hetero_cifar100_iid8}
\end{table*}

\section{Results of AdaptCL}
\label{appendix:performance}
In this section, we show the detailed results of AdaptCL for different datasets with different heterogeneity.

{\bf Supplementary analysis}. As we can see from Fig. \ref{figure:accround}, in addition to our approach, FedAVG-S achieves the highest accuracy due to its ability to take full advantage of all the data compared to the asynchronous approach. In the asynchronous approach, especially the completely asynchronous approach, i.e., FedAsync-S and DC-ASGD, they try to mitigate the gradient staleness issue somehow. However, essentially, the information learned from the data on the slow workers is insufficient, which leads to the knowledge of this part of the data not being well integrated into the global model. In SSP, the server needs to do $W*T$ aggregations (the server only needs to do $T$ aggregations in FedAVG). The worker may encounter the server is updating the model when it submits the model, so it needs to wait and cause the overall time of SSP to increase.

The reason for the slight drop in accuracy of the experiments on the Tiny-ImageNet we consider is related to not pruning the last layer of each resnet block. We will try a better way of cropping later to improve the performance on ResNet model.

{\bf Detailed results}. We use four evaluation metrics for comparison, including reduction of top-1 test accuracy ($\Delta$Acc), total training acceleration (Time), reduction of average parameter sizes of all worker models (Param), and reduction of average FLOPs of all worker models (FLOPs). We report the results of IID CIFAR10, Non-IID CIFAR10, IID CIFAR100 and Non-IID CIFAR100 in Tab. \ref{table:differnet_hetero_cifar10_iid}, Tab. \ref{table:differnet_hetero_cifar10_iid8}, Tab. \ref{table:differnet_hetero_cifar100_iid} and Tab. \ref{table:differnet_hetero_cifar100_iid8}, respectively.

From the results, AdaptCL achieves good performance in various heterogeneous environments and datasets. We can find that the reduction of time and the reduction of parameters are basically the same, which indicates that the reduction of update time in the experimental simulation environment relies mainly on the reduction of communication time.

The performance at maximum bandwidth $B_{max}$ = 5MB outperforms the performance at $B_{max}$ = 30MB. Because the communication time accounts for a larger percentage of the update time when $B_{max}$ = 5MB. Considering that training time is less sensitive to the pruning in our experiments (as shown in Fig.\ref{figure:train time sensitivity}), fewer parameters can be pruned to reduce the update time and resolve the heterogeneity problem when $B_{max}$ = 30MB.

\section{Internal performance of AdaptCL}
\label{appendix:Internal performance}
In this section, we add a description of the mechanism inside AdaptCL. 

\begin{figure}[htbp]
    \centering
    \subfigure[Update time of each round]{
        \label{figure:roundtime}
        \includegraphics[width=.5\linewidth]{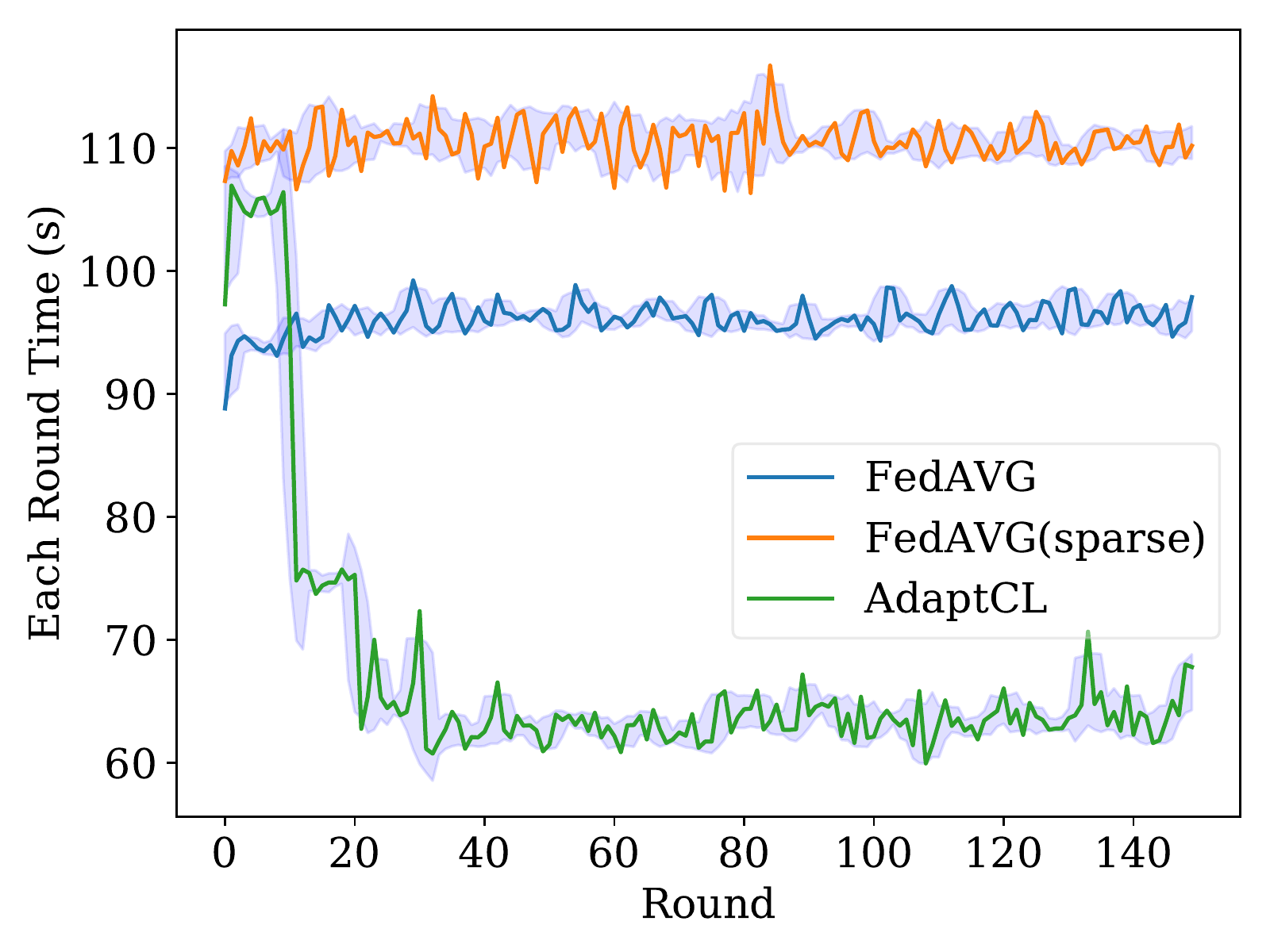}
    }
    \subfigure[Mean update time of partial workers and corresponding heterogeneity]{
    \label{figure:worker_time}
	\includegraphics[width=.5\linewidth]{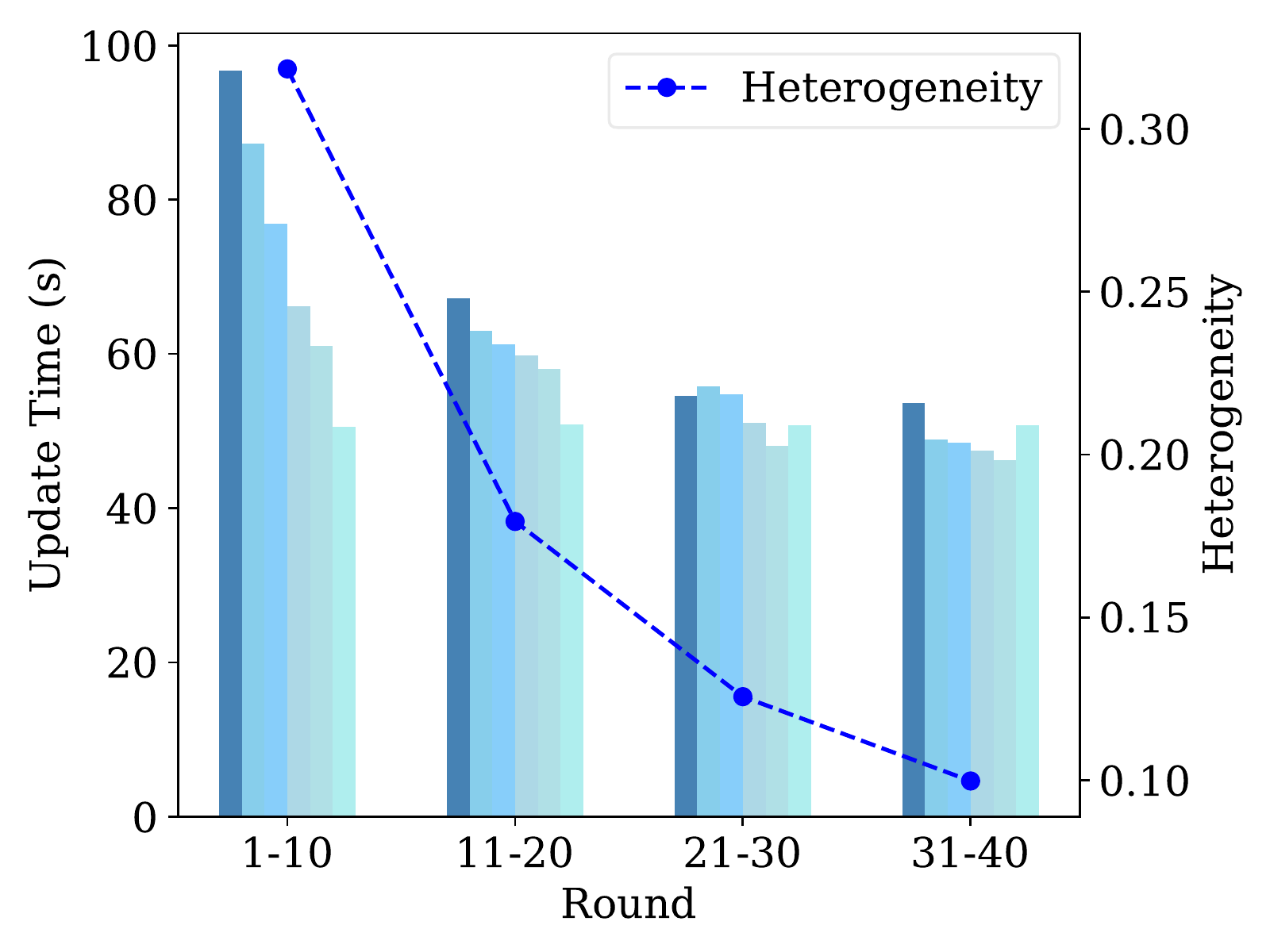}
    }
    \caption{Internal performance of AdaptCL ($H$=0.32, a: CIFAR10, Non-IID, $s$=80, b: CIFAR100, Non-IID, $s$=80).}
    \label{figure:Inter}
\end{figure}

{\bf Internal mechanism of AdaptCL}. The previous experimental results show that AdaptCL outperforms the other methods, and next, we analyze the reasons for the good results from the internal mechanism of AdaptCL. The update time of each round is illustrated in Fig. \ref{figure:roundtime}, where AdaptCL starts with the same amount of time as FedAVG-S and then gradually decreases with subsequent pruning. Finally, the update time of each round stabilizes in early rounds. Moreover, we selected six workers to show their average update time during the first four pruning intervals, as shown in Fig. \ref{figure:worker_time}. As training proceeds, the pruned rate learning algorithm assigns an adaptive pruned rate to each worker, and the update time of all workers gradually tends to the fastest worker. Meanwhile, the heterogeneity of update time between workers rapidly decreases. There are no more draggers in the system by internal adjusting, and the system's efficiency increases dramatically.

\begin{figure}[htbp]
\begin{center}
  \includegraphics[width=0.8\linewidth]{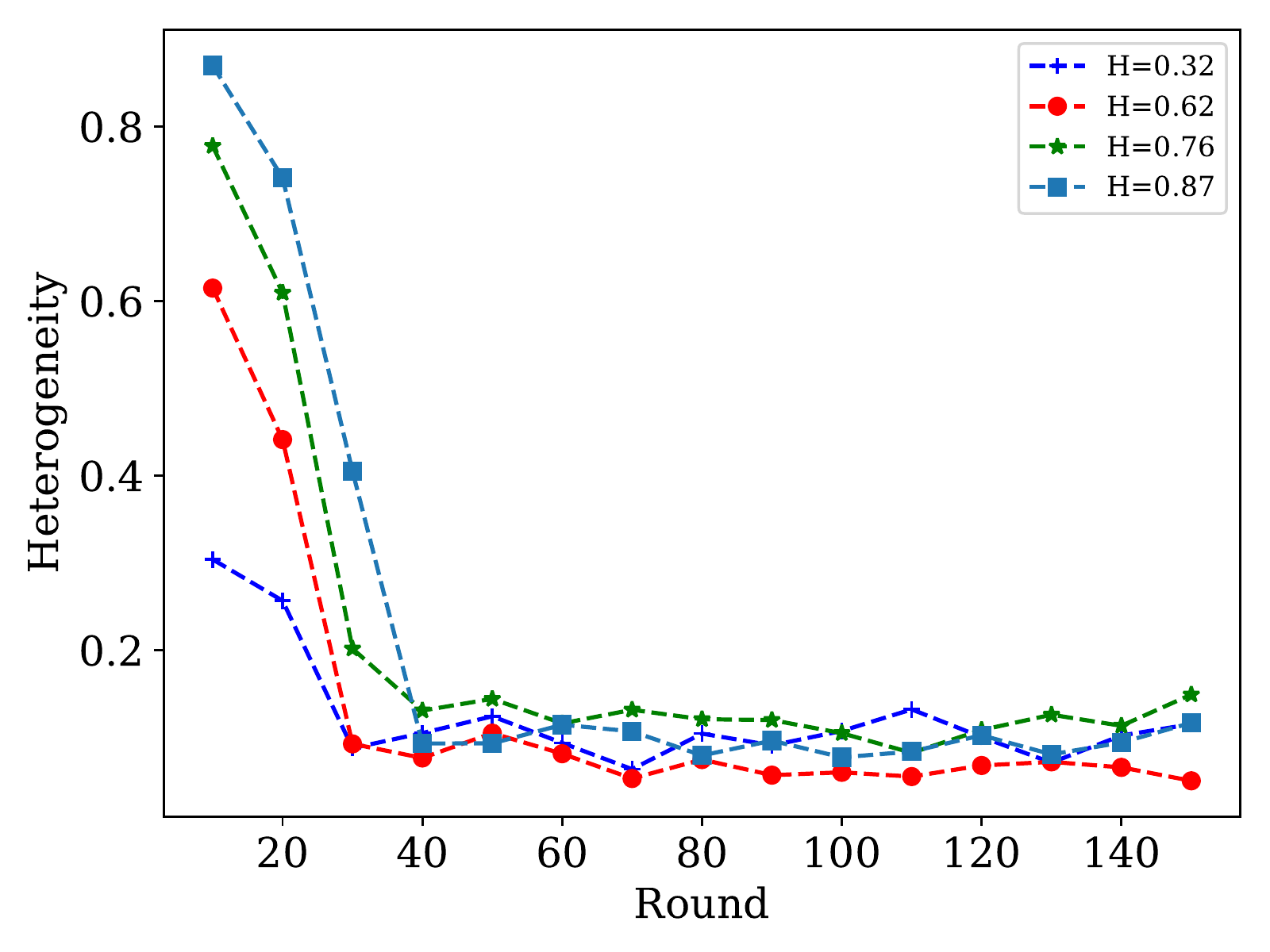}
\end{center}
  \caption{Heterogeneity of update time.}
\label{figure:roundhetero}
\end{figure}

{\bf Environmental heterogeneity}. In AdaptCL, the heterogeneity of update time between workers rapidly decreases and stabilizes quickly regardless of the initial degree of heterogeneity (Fig. \ref{figure:roundhetero}). This reflects that our pruned rate learning algorithm can dynamically give an adaptive pruned rate to make the worker update time converge to the minimum update time.

\begin{figure}[htbp]
\begin{center}
  \includegraphics[width=0.8\linewidth]{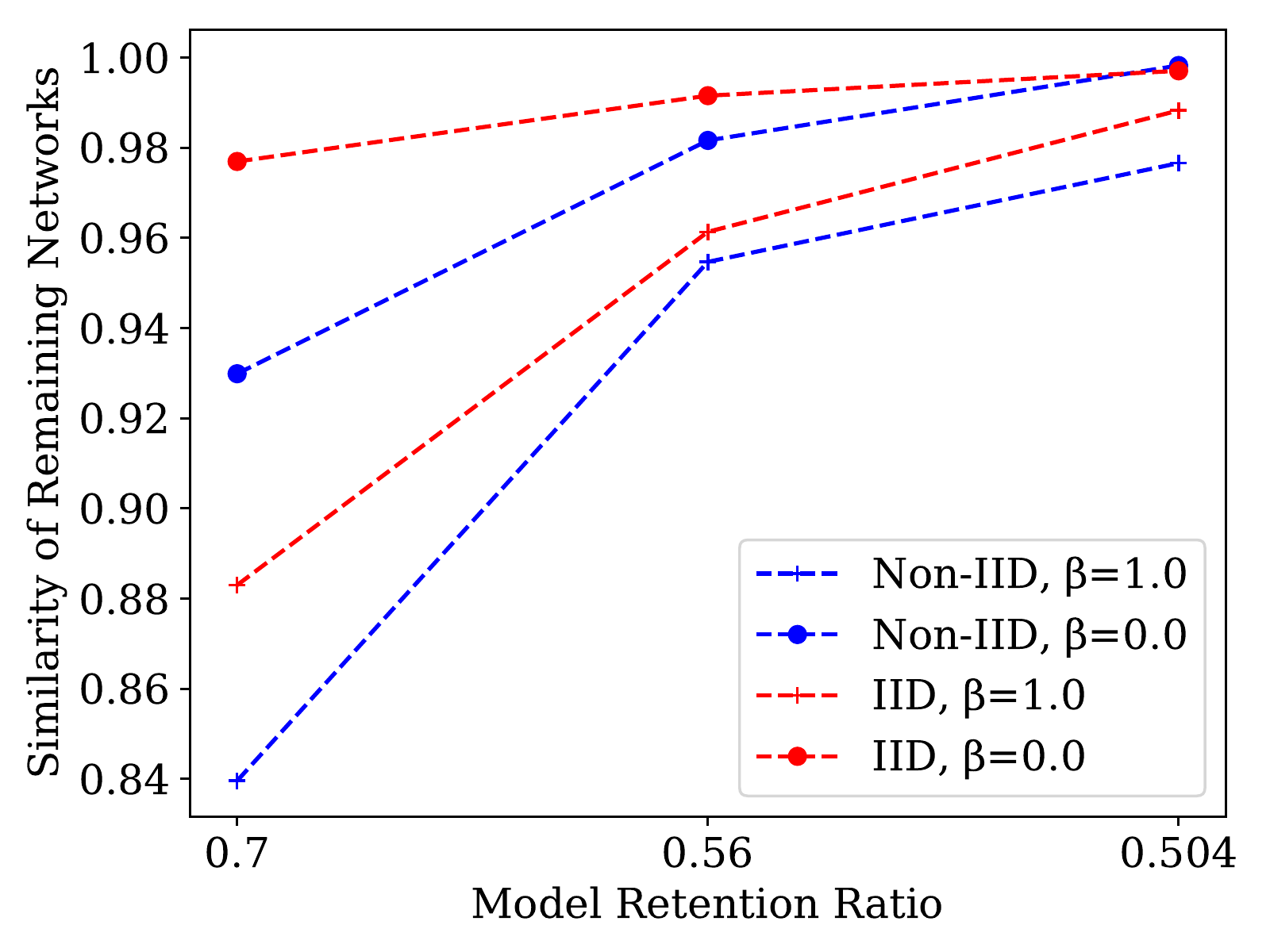}
\end{center}
  \caption{The remaining networks become similar as the pruning proceeds (CIFAR100).}
\label{figure:saved_model}
\end{figure}

{\bf Remaining network}. We choose two workers (worker two and worker four in Tab. \ref{table:pruned rate}) with identical pruned rates per round to compare their remaining networks' similarity. We define the similarity by the mean value of the ratio of the intersection size of units per layer to the union size of units per layer, as in Eq. \ref{equation:similarity}. We do not calculate the similarity of the unpruned layers. We next explore the effect of data distribution and pruning position on the similarity of the remaining network.

As we can see in Fig. \ref{figure:saved_model}, in both the IID and Non-IID cases, the similarity between the remaining networks of the two workers gradually increases as the pruning continues. This means that even though the data types of the workers are different, high-importance units are similar. However, the similarity of the remaining networks in Non-IID is still lower than in the IID case. In both the IID and Non-IId cases, the similarity of the remaining networks is higher when pruning is placed before local training ($\beta$=0.0). Different workers have just received the same global model parameters and are likely to prune the same units. In contrast, the similarity of the remaining networks is lower when pruning is placed after local training ($\beta$=1.0).

\begin{table}[htbp]
\begin{center}
\resizebox{!}{1.5cm}{
\begin{tabular}{cccccc}
 \toprule
 \multirow{2}{*}{Dataset} & \multirow{2}{*}{$PI$} & \multicolumn{2}{c}{IID($s$=0)} &  \multicolumn{2}{c}{Non-IID($s$=80)}\\
   \cmidrule(r) {3-4} \cmidrule(r) {5-6}
 &  & Acc(\%) & Time(min) & Acc(\%) & Time(min) \\
\midrule
\multirow{2}{*}{CIFAR10} 
 &  5  &{\bf 87.38}  & {\bf 153.82}   & {\bf 81.07} & {\bf 155.89}   \\
 &  10 &87.35 &171.80  &80.9 &  157.58  \\

\cmidrule {2 -6}
\multirow{2}{*}{CIFAR100}
 &  5  & 61.23 & 154.07 & {\bf 51.97 } & 156.56  \\
 &  10 & {\bf 62.17} & {\bf 152.45}  &  51.85 & {\bf 154.32} \\
\bottomrule
\end{tabular}}
\end{center}
\caption{Performance of AdaptCL under different pruning intervals.}
\label{table:pruning interval}
\end{table}

{\bf Pruning interval $PI$}. A smaller pruning interval means, on the one hand, that AdaptCL can unify all worker update time at an earlier time, leading to a shorter overall training time, and a pruned model can have more time to recover. On the other hand, a smaller interval between cuts may result in the model being cut again before it recovers, leading to a decrease in accuracy, and the estimated update time of a model is more likely to receive random factors. We report the results of the smaller pruning interval together with the previous results in the Tab. \ref{table:pruning interval}. As we 
can see, better results are achieved with small pruning intervals, i.e., an improvement in both accuracy and time reduction. This indicates that using a small pruning interval is more suitable in the case of small update time fluctuations.

\section{Further Enhancements of AdaptCL}
\label{appendix:Further Enhancements}
In this section we discuss some of the ways in which AdaptCL can be further enhanced.

\begin{figure}[htbp]
\begin{center}
  \includegraphics[width=0.8\linewidth]{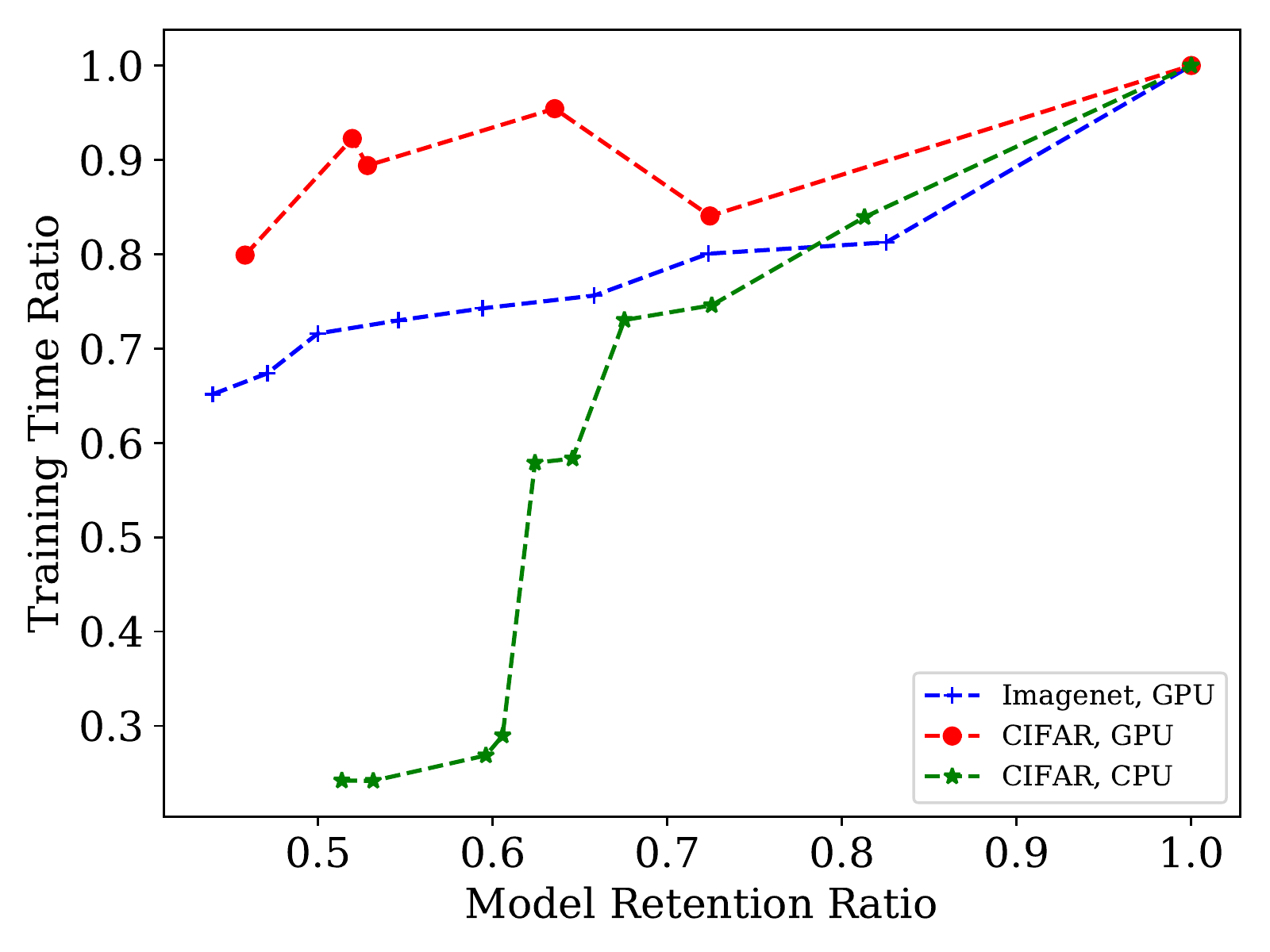}
\end{center}
  \caption{Sensitivity of model training time to pruning on different devices.}
\label{figure:train time sensitivity}
\end{figure}

{\bf Training sensitivity}. As shown in Fig. \ref{figure:train time sensitivity}, the sensitivity of model training time to pruning varies significantly across devices. On GPU, the training time does not change significantly after pruning, which means more parameters need to be pruned to reduce the communication time to achieve the desired update time. However, on the CPU, the training time is more sensitive to pruning. In the same heterogeneous environment, to achieve the same low heterogeneity, the workers running on the CPU do not have to prune many parameters so that the model retention is larger and the accuracy is higher.

\begin{table}[htbp]
\begin{center}
\resizebox{!}{1.1 cm}{
\begin{tabular}{cccc}
 \toprule
 $H$(device) &  Acc(\%) & Param ($\downarrow$) & Minimun Param \\
\midrule
 0.76(GPU)  & 86.78\% & 76.62\% & 6.98\%  \\
 0.62(GPU)  & 87.15\% & 67.56\% & 12.91\%  \\
 0.70(CPU)   & 87.22\% & 45.34\% & 32.98\% \\
\bottomrule
\end{tabular}}
\end{center}
\caption{Experimental results on different device (CIFAR10, IID).}
\label{table:IID,CPU}
\end{table}

\begin{table}[htbp]
\begin{center}
\resizebox{!}{1.1 cm}{
\begin{tabular}{cccc}
 \toprule
 $H$(device) &  Acc(\%) & Param ($\downarrow$) & Minimun Param \\
\midrule
 0.62(GPU)  & 79.92\% & 67.25\% & 11.42\%  \\
 0.32(GPU)  & 80.90\% & 47.65\% & 26.28\%  \\
 0.52(CPU)   & 85.21\% & 41.05\% & 31.63\% \\
\bottomrule
\end{tabular}}
\end{center}
\caption{Experimental results on different device (CIFAR10, Non-IID, $s$=80).}
\label{table:Non-IID,CPU}
\end{table}

We report the results of CIFAR10 training on CPU in Tab. \ref{table:IID,CPU} and Tab. \ref{table:Non-IID,CPU}. As we can see, the workers training on the CPU cut off fewer parameters on average, and the slowest worker's model is much larger than the model of the slowest worker on GPU. Thus the accuracy is also higher on the CPU. Especially, the accuracy is even 4\% higher in the Non-IID case. Because the model of the slowest worker running on GPU is small, resulting in the class data it has not being fully utilized, while the other workers do not have much data in that class.

In more realistic scenarios, where edge workers are often not equipped with chips like GPU, training time is more sensitive to pruning. This means that AdaptCL can work better in realistic scenarios.

\begin{table}[htbp]
\begin{center}
\resizebox{!}{1.3 cm}{
\begin{tabular}{cccc}
 \toprule
 Sparsity &  Acc(\%) & Param Compression & Time (min)\\
\midrule
 0.0  & 80.90\% & 0.00\%  & 157.60\\
 0.7 & 81.49\% & 31.90\% & 150.31\\
 0.9   & 81.95\% & 76.05\% & 140.93\\
 0.99   & 79.55\% & 95.92\% & 136.76\\
\bottomrule
\end{tabular}}
\end{center}
\caption{Experimental results of Adapt+DGC (CIFAR10, Non-IID, $s$=80). Sparsity represents the ratio of uncommitted weights.}
\label{table:dgc}
\end{table}

{\bf Combine with other methods}. Our approach allows different capability workers to take on different amounts of tasks, thus unifying the update time for all workers. As we can see from Fig. \ref{figure:roundtime} and Fig. \ref{figure:roundhetero}, AdaptCL achieves uniformity requires only three or four pruning intervals. After unification, our approach can also be combined with various local improvement methods that aim to accelerate single-round model commitment (as discussed in Sec. \ref{subsection:Efficient Collaborative Learning}). 

DGC \cite{lin2018deep} reduces the communication overhead by committing only some of the most important gradients, and the uncommitted gradients are accumulated locally until a certain threshold is reached. A series of methods such as \emph{momentum corelation} and \emph{momentum factor masking} are designed to solve the problems caused by uncommitted gradients. We use DGC to compress weights and report the AdaptCL+DGC results in Tab. \ref{table:dgc}. As we can see, moderate weight compression not only shortens the training time but also brings an accuracy improvement (e.g., 1\% accuracy improvement with 10.58\% time saving when the compression is 76.05\%). 

This means that our approach can address the global cause (as discussed in Sec. \ref{subsection:Efficient Collaborative Learning}) that affects the efficiency of collaborative learning while being compatible with other ways to address the local cause, leading to more efficient collaborative learning.

\end{appendices}

\end{document}